# Avoiding and Escaping Depressions
# in Real-Time Heuristic Search

**Carlos Hernández**                                    CHERNAN@UCSC.CL
*Departamento de Ingeniería Informática*
*Universidad Católica de la Santísima Concepción*
*Caupolicán 491, Concepción, Chile*

**Jorge A. Baier**                                       JABAIER@ING.PUC.CL
*Departamento de Ciencia de la Computación*
*Pontificia Universidad Católica de Chile*
*Vicuña Mackenna 4860, Santiago, Chile*

## Abstract

Heuristics used for solving hard real-time search problems have regions with depressions. Such regions are bounded areas of the search space in which the heuristic function is inaccurate compared to the actual cost to reach a solution. Early real-time search algorithms, like LRTA*, easily become trapped in those regions since the heuristic values of their states may need to be updated multiple times, which results in costly solutions. State-of-the-art real-time search algorithms, like LSS-LRTA* or LRTA*(k), improve LRTA*'s mechanism to update the heuristic, resulting in improved performance. Those algorithms, however, do not guide search towards avoiding depressed regions. This paper presents *depression avoidance*, a simple real-time search principle to guide search towards avoiding states that have been marked as part of a heuristic depression. We propose two ways in which depression avoidance can be implemented: *mark-and-avoid* and *move-to-border*. We implement these strategies on top of LSS-LRTA* and RTAA*, producing 4 new real-time heuristic search algorithms: aLSS-LRTA*, daLSS-LRTA*, aRTAA*, and daRTAA*. When the objective is to find a single solution by running the real-time search algorithm once, we show that daLSS-LRTA* and daRTAA* outperform their predecessors sometimes by one order of magnitude. Of the four new algorithms, daRTAA* produces the best solutions given a fixed deadline on the average time allowed per planning episode. We prove all our algorithms have good theoretical properties: in finite search spaces, they find a solution if one exists, and converge to an optimal after a number of trials.

## 1. Introduction

Many real-world applications require agents to act quickly in a possibly unknown environment. Such is the case, for example, of autonomous robots or vehicles moving quickly through initially unknown terrain (Koenig, 2001). It is also the case of virtual agents in games (e.g., Warcraft, Starcraft), in which the time dedicated by the game software to perform tasks such as path-finding for *all* virtual agents is very limited. Actually, companies impose limits on the order of 1 millisecond to perform these tasks (Bulitko, Björnsson, Sturtevant, & Lawrence, 2011). Therefore, there is usually no time to plan for full trajectories in advance; rather, path-finding has to be carried out in a real-time fashion.

Real-time search (e.g., Korf, 1990; Weiss, 1999; Edelkamp & Schrödl, 2011) is a standard paradigm for solving search problems in which the environment is not fully known in advance and agents have to act quickly. Instead of running a computationally expensive procedure to generate a conditional plan at the outset, real-time algorithms interleave planning and execution. As such, they usually run a computationally inexpensive lookahead-update-act cycle, in which search is carried out to select the next move (lookahead phase), then learning is carried out (update phase), and finally an action is executed which may involve observing the environment (act phase). Like standard A* search (Hart, Nilsson, & Raphael, 1968), they use a heuristic function to guide action selection. As the environment is unveiled, the algorithm updates its internal belief about the structure of the search space, updating (i.e. *learning*) the heuristic value for some states. The lookahead-update-act cycle is executed until a solution is found.

Early heuristic real-time algorithms like *Learning Real-Time A** (LRTA*) and *Real-Time A** (RTA*) (Korf, 1990) are amenable for settings in which the environment is initially unknown. These algorithms will perform poorly in the presence of *heuristic depressions* (Ishida, 1992). Intuitively, a heuristic depression is a bounded region of the search space in which the heuristic is inaccurate with respect to the heuristic values of the states in the border of the region. When an agent controlled by LRTA* or RTA* enters a region of the search space that conforms a heuristic depression it will usually become "trapped". In order to leave the heuristically depressed region, the agent will need to visit and update many states in this region, potentially several times. Furthermore, in many applications, such as games, the behavior of the agent in a depression may look irrational and thus it is undesirable.

State-of-the-art heuristic real-time search algorithms that are suitable for applications with initially unknown environments are capable of escaping heuristic depressions more quickly than LRTA* or RTA*. They do so by performing more lookahead search, more learning, or a combination of both. More search involves selecting an action by looking farther away in the search space. More learning usually involves updating the heuristic of several states in a single iteration. There are many algorithms that use one or a combination of these techniques (e.g., Hernández & Meseguer, 2005; Bulitko & Lee, 2006; Koenig & Likhachev, 2006b; Hernández & Meseguer, 2007; Rayner, Davison, Bulitko, Anderson, & Lu, 2007; Björnsson, Bulitko, & Sturtevant, 2009; Koenig & Sun, 2009). As a result, these algorithms perform better than LRTA*, spending fewer moves trapped in depressions.

Two algorithms representative of the state of the art in real-time search for initially unknown environments are LSS-LRTA* (Koenig & Sun, 2009) and RTAA* (Koenig & Likhachev, 2006a). These algorithms generalize LRTA* by performing more search and more learning in each episode. Both algorithms have been shown to perform very well in practice. However, despite the use of more elaborate techniques, they may still perform poorly in the presence of heuristic depressions. This is because they may sometimes rely on increasing the heuristic value of states inside the depressions as a mechanism to exit them.

In this paper we study techniques that allow us to improve the performance of real-time search algorithms by making them explicitly aware of heuristic depressions, and then by guiding the search in order to avoid and, therefore, escape depressions. Specifically, the contributions of this paper are as follows.

- We provide new empirical evidence that shows that RTAA* outperforms LSS-LRTA* in game map benchmarks in the first trial, which means that whenever there is a single chance to run one of those real-time heuristic search algorithms to solve a search problem, RTAA* finds better solutions than LSS-LRTA* while making the same search effort. Before, Koenig and Likhachev (2006b) had shown similar performance results but in mazes. This is important since LSS-LRTA*, and not RTAA*, is the algorithm that has received more attention by the real-time heuristic search community. In this paper we consider incorporating our techniques to both LSS-LRTA* and RTAA*.

- We propose a definition for *cost-sensitive heuristic depressions*, which is a more general notion than Ishida's (1992) notion of heuristic depression since it incorporates action costs. We illustrate that our depressions better describe the regions of the search space in which real-time search algorithms get trapped.

- We propose a simple principle to *actively* guide search towards avoiding cost-sensitive heuristic depressions that we call *depression avoidance*, together with two strategies to implement depression avoidance which can be incorporated into state-of-the-art real-time heuristic search algorithms: *mark-and-avoid* and *move-to-border*.

- We propose four new real-time search algorithms; two based on mark-and-avoid, aLSS-LRTA*, aRTAA*, and two based on move-to-border: daLSS-LRTA*, and daRTAA*. The algorithms are the result of implementing depression avoidance on top of RTAA* and LSS-LRTA*.

- We prove that all our algorithms have desirable properties: heuristic consistency is preserved, they terminate if a solution exists, and they eventually converge to an optimal solution after running a sufficiently large, finite number of trials.

- We carry out an extensive empirical evaluation of our algorithms over deployed game benchmarks and mazes. Our evaluation shows that our algorithms outperform existing algorithms in both game maps and mazes. When little time is allowed for the lookahead phase, two of our algorithms, daLSS-LRTA* and daRTAA*, outperform existing ones by an order of magnitude.

Some of the contributions of this paper have been published in conference papers (Hernández & Baier, 2011d, 2011c). This article includes new material that has not been presented before. In particular:

- We describe and evaluate daLSS-LRTA*, an algorithm that is presented in this article for the first time.

- We include full proofs for the termination results (Theorem 6), and a new theoretical result (Theorem 7) on the convergence of all our algorithms.

- We extend previously published empirical results by including maze benchmarks, which had not been previously considered, and by including more game domains and problems.

- Finally, we discuss in detail some scenarios at which our techniques may not perform particularly good.

The rest of the paper is organized as follows. In Section 2 we explain basic concepts of real-time search. We continue presenting LSS-LRTA* and RTAA*, and we extend the results available in the literature by comparing them over game maps. We continue elaborating on the concept of heuristic depression. We then describe our strategies for implementing depression avoidance and the algorithms that result from applying each of them to LSS-LRTA* and RTAA*. We continue with a detailed theoretical and experimental analysis. Then, we present a discussion of our approach and evaluation. We finish with a summary.

## 2. Preliminaries

A search problem $P$ is a tuple $(S, A, c, s_0, G)$, where $(S, A)$ is a digraph that represents the search space. The set $S$ represents the *states* and the arcs in $A$ represent all available actions. $A$ does not contain elements of form $(x, x)$. In addition, the cost function $c : A \mapsto \mathbb{R}^+$ associates a cost to each of the available actions. Finally, $s_0 \in S$ is the start state, and $G \subseteq S$ is a set of goal states. In this paper we assume search spaces are undirected; i.e., whenever $(u, v)$ is in $A$, then so is $(v, u)$. Furthermore, $c(u, v) = c(v, u)$, for all $(u, v) \in A$. The successors of a state $u$ are defined by $Succ(u) = \{v \mid (u, v) \in A\}$. Two states are *neighbors* if they are successors of each other.

A heuristic function $h : S \mapsto [0, \infty)$ associates to each state $s$ an approximation $h(s)$ of the cost of a path from $s$ to a goal state. We denote by $h^*(s)$ the cost of an optimal path to reach a solution from $s$.

A heuristic $h$ is *consistent* if and only if $h(g) = 0$ for all $g \in G$ and $h(s) \leq c(s, s') + h(s')$ for all states $s' \in Succ(s)$. If $h$ is consistent and $C(s, s')$ is the cost of any path between two states $s$ and $s'$, then $h(s) \leq C(s, s') + h(s')$. Furthermore, if $h$ is consistent it is easy to prove that it is also *admissible*; i.e., $h(s)$ underestimates $h^*(s)$. For more details on these definitions, we refer the reader to the book authored by Pearl (1984).

We refer to $h(s)$ as the *h*-value of $s$ and assume familiarity with the A* algorithm (Hart et al., 1968): $g(s)$ denotes the cost of the path from the start state to $s$, and $f(s)$ is defined as $g(s) + h(s)$. The *f*-value and *g*-value of $s$ refer to $f(s)$ and $g(s)$ respectively.

### 2.1 Real-Time Search

The objective of a real-time search algorithm is to make an agent travel from an initial state to a goal state performing, between moves, an amount of computation bounded by a constant. An example situation is path-finding in a priori unknown grid-like environments. There the agent has sufficient memory to store its current belief about the structure of the search space. In addition, the *free-space assumption* (Zelinsky, 1992; Koenig, Tovey, & Smirnov, 2003) is taken: the environment is initially assumed as obstacle-free. The agent is capable of a limited form of sensing: only obstacles in the neighbor states can be detected. When obstacles are detected, the agent updates its map accordingly.

Many state-of-the-art real-time heuristic search algorithms can be described by the pseudo-code in Algorithm 1. The algorithm iteratively executes a lookahead-update-act cycle until the goal is reached. The lookahead phase (Line 4–6) determines the next state to move to, the update phase (Line 7) updates the heuristic, and the act phase (Line 8) moves the agent to its next position. The lookahead-update part of the cycle (Lines 4–7) is referred to as the *planning episode* throughout the paper.

---

**Algorithm 1:** A generic real-time heuristic search algorithm

---

**Input**: A search problem $P$, and a heuristic function $h$.
**Side Effect**: The agent is moved from the initial state to a goal state if a trajectory exists

**1** $h_0 \leftarrow h$
**2** $s_{current} \leftarrow s_0$
**3** **while** $s_{current} \notin G$ **do**
**4**      LookAhead ()
**5**      **if** $Open = \emptyset$ **then return** no-solution
**6**      $s_{next} \leftarrow$ Extract-Best-State()
**7**      Update ()
**8**      move the agent from $s_{current}$ to $s_{next}$ through the path identified by LookAhead. Stop if an action cost along the path is updated.
**9**      $s_{current} \leftarrow$ current agent position
**10**      update action costs (if they have increased)

---

The generic algorithm has three local variables: $s_{current}$ stores the current position of the agent, $c(s, s')$ contains the cost of moving from state $s$ to a successor $s'$, and $h$ is such that $h(s)$ contains the heuristic value for $s$. All three variables may change over time. In path-finding tasks, when the environment is initially unknown, the initial value of $c$ is such that no obstacles are assumed; i.e., $c(s, s') < \infty$ for any two neighbor states $s, s'$. The initial value of $h(s)$, for every $s$, is given as a parameter.

The generic algorithm receives as input a search problem $P$, and starts off by initializing some useful variables (Lines 1–2). In $h_0$ it records the initial value of $h$, for all states in $P$, and in $s_{current}$ it stores the initial position of the agent, $s_0$. We assume the cost of an arc cannot decrease. In particular, arc costs increase to infinity when an obstacle is discovered.

In the lookahead phase (Lines 4–6), the algorithm determines where to proceed next. The Lookahead() procedure in Line 4 implements a bounded search procedure that expands states from the current state $s_{current}$. The set of states generated by this call is referred to as *local search space*. Different choices can be made to implement this procedure. *Real-Time A\** (RTA\*) and *Learning Real-Time A\** (LRTA\*)—two early algorithms proposed by Korf (1990) —and other modern real-time search algorithms run a search from the current state up to a fixed depth (e.g., Bulitko & Lee, 2006). Another common option is to run a bounded A\* search; such a choice is taken by *Local Search Space LRTA\** (LSS-LRTA\*) (Koenig & Sun, 2009), and *Real-Time Adaptive A\** (RTAA\*) (Koenig & Likhachev, 2006b). Algorithm 2 shows the pseudo-code for bounded A\*. Note that at most $k$ states are expanded, where $k$ is a parameter of the algorithm usually referred to as the *lookahead parameter*. The pseudo code of the generic real-time search algorithm assumes that the call to Lookahead() stores the *frontier* of the local search space in *Open*, and, moreover, that if a goal state is found during search, such a state is not removed from the frontier (in the bounded A\* pseudo-code this is guaranteed by the condition in Line 7).

In the last step of the lookahead phase (Line 6, Algorithm 1), the variable containing the next state to move to, $s_{next}$, is assigned. Here, most algorithms select the state in the search frontier that is estimated to be closest to a goal state. When A\* lookahead is used, such a state usually corresponds to a state with minimum $f$-value in *Open*. Thus A\*-based lookahead algorithms use Algorithm 3 to implement the Extract-Best-State() function.

---

**Algorithm 2:** Bounded A* lookahead

---

1 **procedure A*** ()
2     **for each** $s \in S$ **do** $g(s) \leftarrow \infty$
3     $g(s_{current}) \leftarrow 0$
4     $Open \leftarrow \emptyset$
5     Insert $s_{current}$ into $Open$
6     $expansions \leftarrow 0$
7     **while** *each $s' \in Open$ with minimum $f$-value is such that $s' \notin G$ and expansions $< k$* **do**
8         Remove state $s$ with smallest $f$-value from $Open$
9         Insert $s$ into $Closed$
10        **for each** $s' \in Succ(s)$ **do**
11            **if** $g(s') > g(s) + c(s, s')$ **then**
12                $g(s') \leftarrow g(s) + c(s, s')$
13                $s'.\text{back} = s$
14                **if** $s' \in Open$ **then** remove $s'$ from $Open$
15                Insert $s'$ in $Open$
16        $expansions \leftarrow expansions + 1$

---

**Algorithm 3:** Selection of the Best State used by LSS-LRTA*, RTAA*, and other algorithms.

---

1 **procedure Extract-Best-State** ()
2     **return** $\text{argmin}_{s' \in Open} g(s') + h(s')$

---

In the update phase (Line 7, Algorithm 1), the heuristic of some states in the search space is updated to a value that is a better estimate of the true cost to reach a solution, while staying consistent. After exploring states in the vicinity of $s_{current}$, the algorithm gains information about the heuristic value of a number of states. Using this information, the $h$-value of $s_{current}$—and potentially that of other states in the search space—can be updated in such a way that they reflect a better estimation of the cost to reach a solution. Since after the update the heuristic of some states are updated to a value closer to the true cost, this phase is also referred to as the *learning phase*.

The literature describes several ways in which one can implement the update of the heuristic, e.g., mini-min (e.g., Korf, 1990), max of mins (Bulitko, 2004), and heuristic bounded propagation (Hernández & Meseguer, 2005). The learning rules that are most relevant to this paper, however, are those implemented by LSS-LRTA* and RTAA*. They are described in detail in the following subsections.

Finally, after learning, the agent attempts to move to the state selected by the Extract-Best-State() function, $s_{next}$. In most implementations, the path to the selected state has been computed already by the Lookahead() procedure (in the case of Algorithm 2, the path is reconstructed using the back pointer that is set in Line 13). When the environment is known in advance, the agent can always move to the destination. However, when the environment is not known in advance, this process can fail (in path-finding, this can occur due to the discovery of an obstacle). If such an obstacle is found, we assume the agent stops moving as soon as it has detected an obstacle. In such cases, the algorithm will update its memory regarding the environment, which typically involves updating the cost function. In our pseudo-code, this is reflected in Line 10.

## 3. LSS-LRTA* and RTAA*

Now we describe LSS-LRTA* and RTAA*, the two state-of-the-art real-time heuristic search algorithms that are most relevant to this paper. We make two small contributions to the understanding of these two algorithms. First, we do an experimental comparison of them over benchmarks that had not been considered before. Second, we prove two theoretical results that aim at understanding the differences between their update mechanisms (Propositions 1 and 2). To our knowledge, none of these results appear in the literature.

### 3.1 LSS-LRTA*

*Local search space LRTA*\** (LSS-LRTA*) was first introduced by Koenig (2004), and later presented in detail by Koenig and Sun (2009). It is an instance of Algorithm 1. Its lookahead procedure is a bounded A* search (Algorithm 2). The next state to move to corresponds to a state in *Open* with the lowest *f*-value; i.e., it uses Algorithm 3 to implement `Extract-Best-State()`.

LSS-LRTA* updates the values of each state *s* in the local search space in such a way that *h(s)* is assigned the *maximum* possible value that guarantees consistency with the states in *Open*. It does so by implementing the `Update()` procedure as a modified Dijkstra's algorithm (Algorithm 4). Since the value of *h* is raised to the maximum, the update mechanism of LSS-LRTA* makes *h* as informed as it can get given the current knowledge about the search space, while maintaining consistency.

---

**Algorithm 4:** LSS-LRTA*'s Modified Dijkstra's Procedure. We assume *Open* list is a queue ordered by *h*-value.

**1 procedure ModifiedDijkstra ()**
**2**   **for each** *state s in Closed* **do**   $h(s) \leftarrow \infty$
**3**   **while** *Closed* $\neq \emptyset$ **do**
**4**    Extract an *s* with minimum *h*-value from *Open*
**5**    **if** $s \in Closed$ **then** delete *s* from *Closed*
**6**    **for each** *s' such that* $s \in Succ(s')$ **do**
**7**     **if** $s' \in Closed$ *and* $h(s') > c(s', s) + h(s)$ **then**
**8**      $h(s') \leftarrow c(s', s) + h(s)$
**9**      **if** $s' \notin Open$ **then** Insert *s'* in *Open*

---

**Algorithm 5:** RTAA*'s Update Procedure

**1 procedure Update ()**
**2**   $f \leftarrow \min_{s \in Open} g(s) + h(s)$
**3**   **for each** $s \in Closed$ **do**
**4**    $h(s) \leftarrow f - g(s)$

---

## 3.2 RTAA*

*Real-Time Adaptive A** (RTAA*) was proposed by Koenig and Likhachev (2006b). It is an instance of Algorithm 1. Its lookahead phase is identical to that of LSS-LRTA*: a bounded A* followed by selecting a state with the lowest $f$-value in *Open* as the next state to move to. However it uses a simpler learning mechanism based on the update rule of the incremental A* search algorithm *Adaptive A** (Koenig & Likhachev, 2006a). Thus, it updates the heuristic value of states in the interior of the local search space (i.e., those stored in A*'s variable *Closed*) using the $f$-value of the best state in *Open*. The procedure is shown in Algorithm 5.

RTAA*'s update procedure is considerably faster in practice than that of LSS-LRTA*. Obtaining the lowest $f$-value of a state in *Open* can be done in constant time if A* is implemented with binary heaps. After that, the algorithm simply iterates through the states in *Closed*. The worst-case performance is then $\mathcal{O}(|Closed|)$. On the other hand, LSS-LRTA*'s update procedure first needs to convert *Open* into a priority queue ordered by $h$ and then may, in the worst case, need to extract $|Open| + |Closed|$ elements from a binary heap. In addition, it expands each node that is ever extracted from the priority queue. The time to complete these operations, in the worst case is $T_{exp} \cdot N + T_b \cdot N \log N$, where $N = |Open| + |Closed|$, $T_{exp}$ is the time taken per expansion, and $T_b$ is a constant factor associated to extraction from the binary heap. The worst-case asymptotic complexity of extraction is thus $\mathcal{O}(N \log N)$. However, since we usually deal with a small $N$ it may be the case that the term $T_{exp} \cdot N$ dominates the expression for time.

We will prove that the heuristic values that RTAA* learns may be less accurate than those of LSS-LRTA*. To state this formally, we introduce some notation. Let $h_n$, for $n > 0$, denote the value of the $h$ variable at the start of iteration $n$ of the main algorithm, or, equivalently, right after the update phase of iteration $n - 1$. We will also denote the heuristic function given as input as $h_0$. Let $k_n(s, s')$ denote the cost of an optimal path from $s$ to $s'$ that traverses states only in *Closed* before ending in $s'$.

**Proposition 1** *Let $s$ be a state in Closed right after the call to A* has returned in the n-th iteration of LSS-LRTA*. Then,*

$$h_{n+1}(s) = \min_{s_b \in Open} k_n(s, s_b) + h_n(s_b). \tag{1}$$

**Proof:** We will show that the value $h(s)$ computed by the modified Dijkstra algorithm for each state $s$ corresponds to the minimum cost of reaching node $s$ from a certain state in a particular graph $\mathcal{G}$. The modified Dijkstra procedure can be seen as a run of the standard Dijkstra algorithm (e.g., Cormen, Leiserson, Rivest, & Stein, 2001) on such a graph.

First we observe that our procedure differs from the standard Dijkstra algorithm in that a non-singleton set of states, namely those in *Open*, are initialized with a finite value for $h$. In the standard Dijkstra algorithm, on the other hand, only the source node is initialized with a cumulative cost of 0 whereas the remaining nodes are initialized to $\infty$. With the facts above in mind, it is straightforward to see that a run of the modified Dijkstra can be interpreted as a run of the *standard* Dijkstra algorithm from node $s_{start}$ of a directed graph $\mathcal{G}$ that is such that:

- Its nodes are exactly those in *Open* ∪ *Closed* plus a distinguished node $s_{start}$.

- It contains an arc $(u, v)$ with cost $c$ if there is an arc $(v, u)$ with cost $c$ in the search graph of $P$ such that one of $v$ or $u$ is not in $Open$.

- It contains an arc of the form $(s_{start}, s)$ with cost $h(s)$ for each $s$ in $Open$.

- It contains no other arcs.

After running the Dijkstra algorithm from $s_{start}$ over $\mathcal{G}$, we obtain, for each node $s$ in $\mathcal{G}$ the cost of an optimal path from $s_{start}$ to $s$. If we interpret such a cost as $h(s)$, for each $s$, Equation 1 holds, which finishes the proof. □

For RTAA* we can prove a sightly different result.

**Proposition 2** *Right after the call to A\* returns in the n-th iteration of RTAA\*, let $s^*$ be the state with lowest f-value in Open, and let $s$ be a state in Closed. Then,*

$$h_{n+1}(s) \leq \min_{s_b \in Open} k_n(s, s_b) + h_n(s_b). \tag{2}$$

*However, if $h_n$ is consistent and $s$ is in the path found by A\* from $s_{current}$ to $s^*$, then*

$$h_{n+1}(s) = \min_{s_b \in Open} k_n(s, s_b) + h_n(s_b). \tag{3}$$

**Proof:** For (2), we use the fact that if the heuristic is consistent, it remains consistent after each RTAA* iteration (a fact proven by Koenig & Likhachev, 2006a), to write the inequality $h_{n+1}(s) \leq \min_{s_b \in Open} k_n(s, s_b) + h_{n+1}(s_b)$. Now note that for every state $s_b$ in $Open$ it holds that $h_n(s) = h_{n+1}(s)$, since the heuristic values of states in $Open$ are not updated. Substituting $h_{n+1}(s)$ in the inequality, we obtain the required result.

For (3), we use the fact proven by Hart et al. (1968) about A*: if consistent heuristics are used, $g(s)$ contains the cost of the cheapest path from the start state to $s$ right after $s$ is extracted from $Open$ (Line 8 in Algorithm 2).

Because A* is run with a consistent heuristic, for any state $s'$ along the (optimal) path found by A* from $s_{current}$ to $s^*$,

$$g(s') = k_n(s_{current}, s'), \text{ and} \tag{4}$$
$$g(s^*) = k_n(s_{current}, s') + k_n(s', s^*). \tag{5}$$

RTAA*'s update rule states that:

$$h_{n+1}(s') = f(s^*) - g(s') = h_n(s^*) + g(s^*) - g(s') \tag{6}$$

Substituting with (4) and (5) in (6), we obtain $h_{n+1}(s') = k_n(s', s^*) + h_n(s^*)$. Finally, observe that

$$k_n(s', s^*) + h(s^*) = \min_{s_b \in Open} k_n(s', s_b) + h_n(s_b).$$

Indeed, if there were an $s^-$ in $Open$ such that $k_n(s', s^*) + h(s^*) > k_n(s', s^-) + h_n(s^-)$, then by adding $g(s')$ to both sides of the inequality, we would have that $f(s^*) > f(s^-)$, which contradicts the fact that $s^*$ is the state with lowest f-value in $Open$. We conclude henceforth that $h_{n+1}(s') = \min_{s_b \in Open} k_n(s', s_b) + h_n(s_b)$. This finishes the proof. □

Proposition 2 implies that, when using consistent heuristics, RTAA*'s update may yield less informed $h$-values than those of LSS-LRTA*. However, at least for some of the states in the local search space, the final $h$-values are equal to those of LSS-LRTA*, and hence they are as informed as they can be given the current knowledge about the search space.

Koenig and Likhachev (2006a) show that for a fixed value of the lookahead parameter, the quality of the solutions obtained by LSS-LRTA* are better on average than those obtained by RTAA* in path-finding tasks over mazes. This is due to the fact that LSS-LRTA*'s heuristic is more informed over time than that of RTAA*. However, they also showed that given a fixed time deadline per planning episode, RTAA* yields *better* solutions than LSS-LRTA*. This is essentially due to the fact that RTAA*'s update mechanism is faster: for a fixed deadline, a *higher* lookahead parameter can be used with RTAA* than with LSS-LRTA*.

We extend Koenig and Likhachev's experimental analysis by running a comparison of the two algorithms on game maps. Table 1 shows average results for LSS-LRTA* and RTAA* ran on 12 different game maps. For each map, we generated 500 random test cases. Observe, for example, that if a deadline of 0.0364 milliseconds is imposed per planning episode we can choose to run RTAA* with a lookahead $k = 128$, whereas we can choose to run LSS-LRTA* only with lookahead $k = 64$. With those parameters, RTAA* obtains a solution about 36% cheaper than LSS-LRTA* does. Figure 1 shows average solution cost versus time per episode. The slopes of the curves suggest that the rate at which RTAA* improves solutions is better than that of LSS-LRTA*, as more time per episode is given. In conclusion RTAA* seems superior to LSS-LRTA* when time is actually important. We thus confirm for a wider range of tasks that, when time per episode matters, RTAA* is better than LSS-LRTA*. These findings are important because mazes (for which previous evaluations existed) are problems with a very particular structure, and results over them do not necessarily generalize to other types of problems.

Although we conclude that RTAA* is an algorithm superior to LSS-LRTA* when it comes to finding a good solution quickly, it is interesting to note that recent research on real-time heuristic search is focused mainly on extending or using LSS-LRTA* (see e.g., Bulitko, Björnsson, & Lawrence, 2010; Bond, Widger, Ruml, & Sun, 2010; Hernández & Baier, 2011d; Sturtevant & Bulitko, 2011), while RTAA* is rarely considered. Since LSS-LRTA* seems to be an algorithm under active study by the community, in this paper we apply our techniques to both algorithms.

## 4. Heuristic Depressions

In real-time search problems heuristics usually contain *depressions*. The identification of depressions is central to our algorithm. Intuitively, a heuristic depression is a bounded region of the search space containing states whose heuristic value is too low with respect to the heuristic values of states in the border of the depression. Depressions exist naturally in heuristics used along with real-time heuristic search algorithms. As we have seen above, real-time heuristic algorithms build solutions incrementally, updating the heuristic values associated to certain states as more information is gathered from the environment.

Ishida (1992) gave a constructive definition for heuristic depressions. The construction starts with a node $s$ such that its heuristic value is equal to or less than those of the surrounding states. The region is then extended by adding a state of its border if all states in the resulting region have a heuristic value lower or equal than those of the states in the border. As a result, the heuristic depression $D$ is a maximal connected component of states such that all states in the boundary of $D$ have a heuristic value that is greater than or equal to the heuristic value of any state in $D$.

| | RTAA* | | | | | LSS-LRTA* | | | | |
|---|---|---|---|---|---|---|---|---|---|---|
| $k$ | Avg. Cost | Time/ep | Exp/ep | Per/ep | Time | Avg. Cost | Time/ep | Exp/ep | Per/ep | Time |
| 1 | 1,146,014 | 0.0004 | 1.0 | 6.1 | 447.9 | 1,146,014 | 0.0012 | 8.7 | 14.8 | 1,259.6 |
| 2 | 919,410 | 0.0006 | 2.0 | 9.4 | 475.4 | 625,693 | 0.0020 | 13.7 | 29.3 | 979.4 |
| 4 | 626,623 | 0.0011 | 4.0 | 17.3 | 468.8 | 372,456 | 0.0034 | 21.3 | 54.3 | 818.1 |
| 8 | 363,109 | 0.0021 | 8.0 | 34.1 | 383.7 | 227,526 | 0.0058 | 33.8 | 102.4 | 653.6 |
| 16 | 188,346 | 0.0040 | 16.0 | 70.1 | 269.1 | 127,753 | 0.0102 | 56.1 | 193.5 | 459.9 |
| 32 | 95,494 | 0.0078 | 32.0 | 152.9 | 192.8 | 72,044 | 0.0187 | 98.7 | 397.7 | 345.3 |
| 64 | 48,268 | 0.0159 | 63.9 | 361.3 | 145.7 | 40,359 | 0.0364 | 184.9 | 903.4 | 279.6 |
| 128 | 25,682 | 0.0326 | 126.4 | 932.3 | 125.8 | 22,471 | 0.0750 | 370.1 | 2,338.1 | 258.2 |
| 256 | 13,962 | 0.0647 | 236.8 | 2,351.8 | 125.6 | 12,264 | 0.1534 | 733.6 | 6,003.8 | 272.2 |
| 512 | 7,704 | 0.1078 | 377.6 | 4,616.7 | 131.6 | 7,275 | 0.2620 | 1,207.5 | 11,548.9 | 312.4 |

Table 1: Average results for the 12 game maps. For lookahead value $k$, we report the solution cost per test case (Avg. Cost), and four measures of efficiency: the runtime per planning episode (Time/ep) in milliseconds, the number of cell expansions per planning episode (Exp/ep), the number of heap percolations per planning episode (Per/ep) and the runtime per test case (Time) in milliseconds. All results were obtained using a Linux machine with an Intel Xeon CPU running at 2GHz and 12 GB RAM.

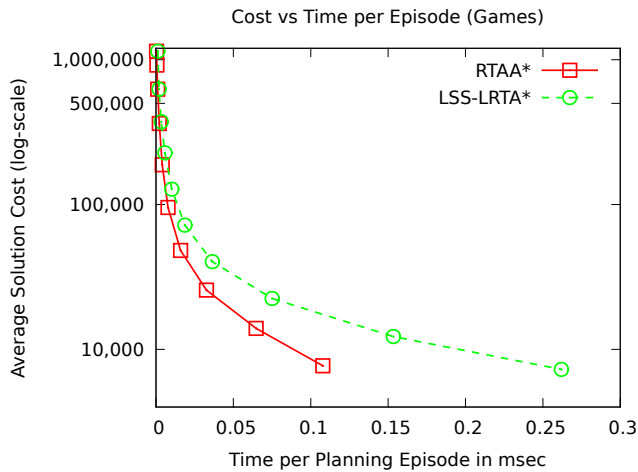

Figure 1: Average solution cost obtained by LSS-LRTA* and RTAA* versus planning time per episode on 12 game maps.

It is known that algorithms like LRTA* behave poorly in the presence of heuristic depressions (Ishida, 1992). To see this, assume that LRTA* is run with lookahead depth equal to 1, such that it only expands the current state, leaving its immediate successors in the search frontier. Assume further that it visits a state in a depression and that the solution node lies outside the depression. To exit the depressed region the agent must follow a path in the interior of the depressed region, say, $s_1 \dots s_n$, finally choosing a state in the border of the region, say $s_e$. While visiting $s_n$, the agent chooses $s_e$ as the next move, which means that $s_e$ *minimizes* the estimated cost to reach a solution among all the neighbors of $s_n$. In problems with uniform action costs, this can only happen if $h(s_e)$ is *lower or equal* than the heuristic value of all other neighbors of $s_n$. This fact actually means that the depression in that region of the search space no longer exists, which can only happen if the heuristic values of states in the originally depressed region have been updated (increased). For LRTA*, the update process may be quite costly: in the worst case *all* states in the depression may need to be updated and each state may need to be updated several times.

Ishida's definition is, nonetheless, restrictive. In fact, it does not take into account the *costs* of the actions needed to move from the interior of the depression to the exterior. A closed region of states may have unrealistically low heuristic values even though the heuristic values in the interior are greater than the ones in the border. We propose a more intuitive notion of depression when costs are taken into account. The formal definition follows.

**Definition 1 (Cost-sensitive heuristic depression)** *A connected component of states $D$ is a cost-sensitive heuristic depression of a heuristic $h$ iff for any state $s \in D$ and every state $s' \notin D$ that is a neighbor of a state in $D$, $h(s) < k(s, s') + h(s')$, where $k(s, s')$ denotes the cost of the cheapest path that starts in $s$, traverses states only in $D$, and ends in $s'$.*

Cost-sensitive heuristic depressions better reflect the regions in which an agent controlled by algorithms such as LRTA* get trapped. To illustrate this, consider the two 4-connected grid-world problems of Figure 2. Gray cells conform an Ishida depression. The union of yellow and gray cells conform a cost-sensitive heuristic depression. Suppose the agent's initial position is the lower-right corner of the Ishida depression (C4 in Figure 2(a), and C7 in Figure 2(b)). Assume further that ties are broken such that the priorities, given from higher to lower, are: down, left, up, and right. For such an initial state, both in situation (a) and situation (b), the agent controlled by LRTA* will visit every state in the *cost-sensitive* heuristic depression before reaching the goal. Indeed, cells in the cost-sensitive depression that are not adjacent to an obstacle are visited exactly 3 times, while cells adjacent to an obstacle are visited 2 times, before the agent escapes the depression, and thus the performance of LRTA* can be described as a linear function on the size of the cost-sensitive depression.

It is interesting to note that for problems like the ones shown in Figure 2, the size of the Ishida depression remains the same while the width of the grid varies. Thus, the size of the Ishida depression is not correlated with the performance of LRTA*. On the other hand, the size of the cost-sensitive heuristic depression *is* a predictor of the cost of the solution

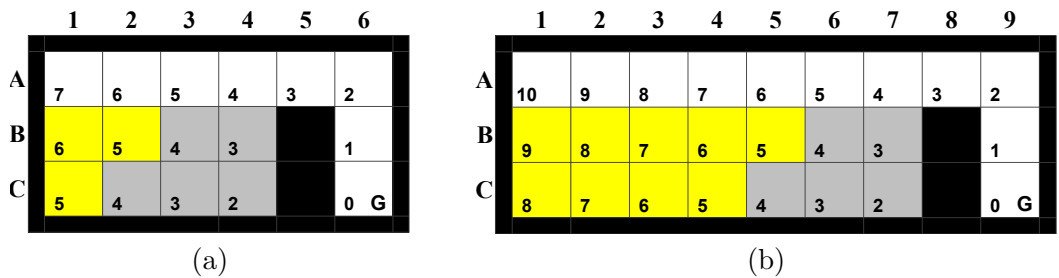

Figure 2: A 4-connected grid-like search space with unitary costs. Black cells are obstacles. The cell with a G is the goal cell. Cells show their $h$-value (Manhattan distance). Ties are broken by giving priority to the down movement, then left, then up, and then right. If the initial position of the agent in situation (a) is C4, then the cells visited by an agent controlled by LRTA* are: C3, C2, C1, B1, B2, B3, B4, C4, C3, B3, C3, C2, B2, C2, C1, B1, C1, B1, B2, B3, B4, A4, A5, A6, B6, and C6. The solution found for (b) is analogous.

## 5. Depression Avoidance

A major issue at solving real-time search problems is the presence of heuristic depressions. State-of-the-art algorithms are able to deal with this problem essentially by doing extensive learning and/or extensive lookahead. By doing more lookahead, chances are that a state outside of a depression is eventually selected to move to. On the other hand, by learning the heuristic values of several states at a time, fewer movements might be needed in order to raise the heuristic values of states in the interior of a depression high enough as to make it disappear. As such, LSS-LRTA*, run with a high value for the lookahead parameter exits the depressions more quickly than LRTA* run with search depth equal to 1 for two reasons: (1) because the heuristic function increases for states in $D$ more quickly and (2) because with a high value for the lookahead parameter it is sometimes possible to escape the depression in one step.

Besides the already discussed LSS-LRTA* and RTAA*, there are many algorithms described in the literature capable of doing extensive lookahead and learning. The lookahead ability of LRTS (Bulitko & Lee, 2006), and TBA* (Björnsson et al., 2009) is parametrized. By using algorithms such as LRTA*($k$) (Hernández & Meseguer, 2005), PLRTA* (Rayner et al., 2007) and LRTA$^*_{LS}$($k$) (Hernández & Meseguer, 2007) one can increase the number of states updated based on a parameter. None of these algorithms however are *aware* of depressions; their design simply allows to escape them because of their ability to do lookahead, learning, or a combination of both. Later, in Section 9, we give a more detailed overview of other related work.

To improve search performance our algorithms avoid depressions, a principle we call *depression avoidance*. Depression avoidance is a simple principle that dictates that search should be guided away from states identified as being in a heuristic depression. There are many ways in which one could conceive the implementation of this principle in a real-time heuristic search algorithm. Below we present two alternative realizations of the principle within the state-of-the-art RTAA* and LSS-LRTA* algorithms. As a result, we propose four new real-time search algorithms, each of which has good theoretical properties.

## 5.1 Depression Avoidance via *Mark-and-Avoid*

This subsection presents a first possible realization of depression avoidance that we call *mark-and-avoid*. With this strategy, we extend the update phase to mark states that we can prove belong to a heuristic depression. We then modify the selection of the best state (i.e., the `Extract-Best-State()` function) to select states that are *not* marked; i.e., states that are not yet proven to be part of a depression.

aLSS-LRTA* is version of LSS-LRTA* that avoids depressions via mark-and-avoid. It is obtained by implementing the `Update()` function using Algorithm 6 and by implementing the `Extract-Best()` function with Algorithm 7. There are two differences between its update procedure and LSS-LRTA*'s. The first is the initialization of the **updated** flag in Lines 2–3. The second is Line 7, which sets $s$.updated to true if the heuristic value for $h$ changes as a result of the update process. In the following section, we formally prove that this means that $s$ was inside a cost-sensitive heuristic depression (Theorem 5).

---

**Algorithm 6:** Modified Dijkstra Procedure used by aLSS-LRTA*.

```
1  procedure ModifiedDijkstra ()
2      if first run then
3        └─ for each s ∈ S do s.updated ← false        /* initialization of update flag */
4      for each s ∈ Closed do h(s) ← ∞
5      while Closed ≠ ∅ do
6          Extract an s with minimum h-value from Open
7          if h(s) > h₀(s) then s.updated = true
8          if s ∈ Closed then delete s from Closed
9          for each s′ such that s ∈ Succ(s′) do
10             if s′ ∈ Closed and h(s′) > c(s′, s) + h(s) then
11                 h(s′) ← c(s′, s) + h(s)
12                 if s′ ∉ Open then Insert s′ in Open
```

---

To select the next state $s_{next}$, aLSS-LRTA* chooses the state with lowest $f$-value from *Open* that has *not* been marked as in a depression. If such a state does not exist, the algorithm selects the state with lowest $f$-value from *Open*, just like LSS-LRTA* would do. Depending on the implementation, the worst-case complexity of this new selection mechanism may be different from that of Algorithm 3. Indeed, if the *Open* list is implemented with a binary heap (as it is our case), the worst-case complexity of Algorithm 7 is $\mathcal{O}(N \log N)$ where $N$ is the size of *Open*. This is because the heap is ordered by $f$-value. On the other hand the worst-case complexity of Algorithm 3 using binary heaps is $\mathcal{O}(1)$. In our experimental results we do not observe, however, a significant degradation in performance due to this factor.

---

**Algorithm 7:** Selection of the next state used by aLSS-LRTA* and aRTAA*

---

**1 function** Extract-Best-State ()

   **2**    **if** *Open contains an s such that s.updated = false* **then**

   **3**       $s \leftarrow \operatorname{argmin}_{s' \in Open \land s'.\text{updated}=false} g(s') + h(s')$

   **4**    **else**

   **5**       $s \leftarrow \operatorname{argmin}_{s' \in Open} g(s') + h(s')$

   **6**    **return** $s$ ;

---

**Example** Figure 3 shows an example that illustrates the difference between LSS-LRTA* and aLSS-LRTA* with the lookahead parameter equal to two. After 4 search episodes, we observe that aLSS-LRTA* avoids the depression, leading the agent to a position that is 2 steps closer to the goal than LSS-LRTA*.

---

**Algorithm 8:** aRTAA*'s Update Procedure

---

**1 procedure** Update ()

   **2**    **if** *first run* **then**

   **3**       **for each** $s \in S$ **do** s.updated $\leftarrow false$      /* initialization of update flag */

   **4**    $f \leftarrow f$-value of the best state in $Open$

   **5**    **for each** $s \in Closed$ **do**

   **6**       $h(s) \leftarrow f - g(s)$

   **7**       **if** $h(s) > h_0(s)$ **then** s.updated $\leftarrow true$

---

With aLSS-LRTA* as a reference, it is straightforward to implement the mark-and-avoid strategy into RTAA*. The update phase of the resulting algorithm, aRTAA*, is just like RTAA*'s but is extended to mark states in a depression (Algorithm 8). The selection of the best state to move to is done in the same way as aLSS-LRTA*, i.e., with Algorithm 7. As a result aRTAA* is a version of RTAA* that aims at avoiding depressions using mark-and-avoid.

## 5.2 Depression Avoidance via *Move-to-Border*

Move-to-border is a more finely grained implementation of depression avoidance. To illustrate the differences, consider that, after lookahead, there is no state $s$ in the frontier of the local search space such that $s$.updated is false. Intuitively, such is a situation in which the agent is "trapped" in a heuristic depression. In this case, aLSS-LRTA* behaves exactly as LRTA* does since all states in the search frontier are marked. Nevertheless, in these cases, we would like the movement of the agent to still be guided *away* from the depression.

In situations in which all states in the frontier of the local search space are already proven as members of a depression, the move-to-border strategy attempts to move to a state that seems *closer to the border* of a depression. As a next state, this strategy chooses the state with best $f$-value among the states whose heuristic has *changed the least*. The intuition behind this behavior is as follows: assume $\Delta(s)$ is the difference between the actual cost to reach a solution from a state $s$ and the initial heuristic value of state $s$. Then, if $s_1$ is a state close to the border of a depression $D$ and $s_2$ is a state farther away from the border and "deep" in the interior of $D$, then $\Delta(s_2) \geq \Delta(s_1)$, because the heuristic of $s_2$ is more imprecise than that of $s_1$. At execution time, $h$ is an estimate of the actual cost to reach a solution.

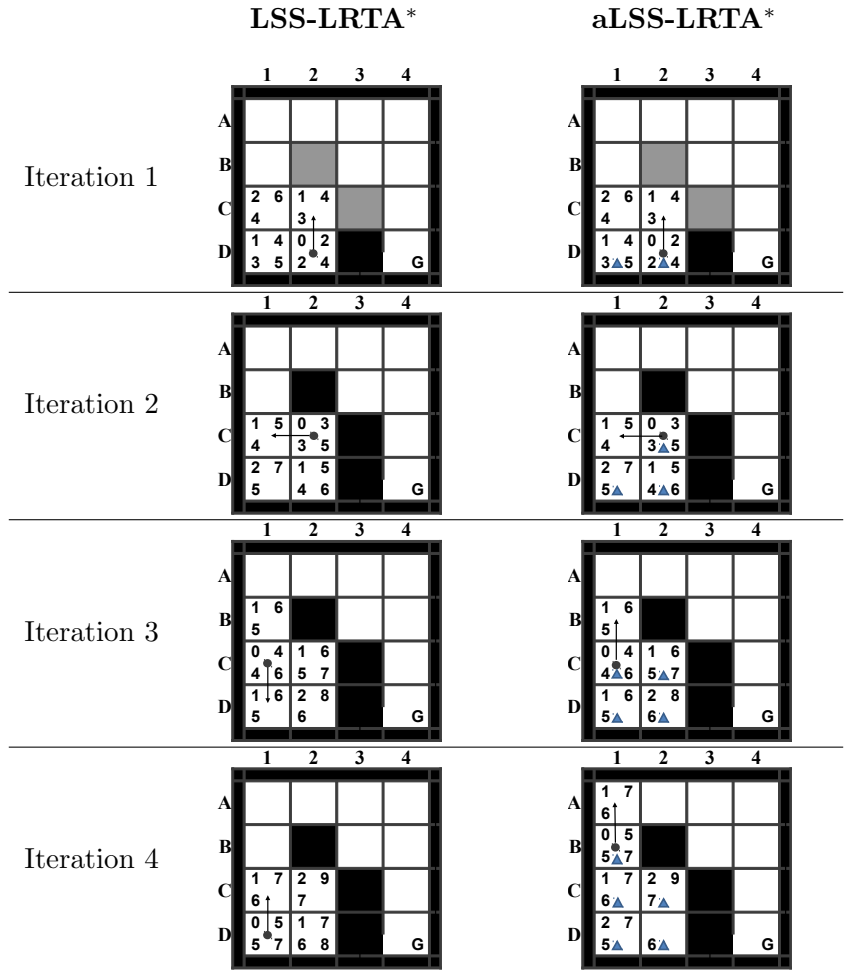

Figure 3: First 4 iterations of LSS-LRTA* (left) and aLSS-LRTA* (right) with lookahead equal to 2 in a 4-connected grid world with unitary action costs, where the initial state is D2, and the goal is D4. Numbers in cell corners denote the $g$-value (upper left), $f$-value (upper right), $h$-value (lower left), and new $h$-value of an expanded cell after an update (lower right). Only cells that have been in a closed list show four numbers. Cells generated but not expanded by A* (i.e., in *Open*) show three numbers, since their $h$-values have not been updated. Triangles (▲) denote states with updated flag set to true after the search episode. The heuristic used is the Manhattan distance. We assume ties are broken by choosing first the right then bottom then the left and then top adjacent cell. The position of the agent is given by the dot. A grid cell is shaded (gray) if it is a blocked cell that the agent has not sensed yet. A grid cell is black if it is a blocked cell that the agent has already sensed. The best state chosen to move the agent to after lookahead search is pointed by an arrow.

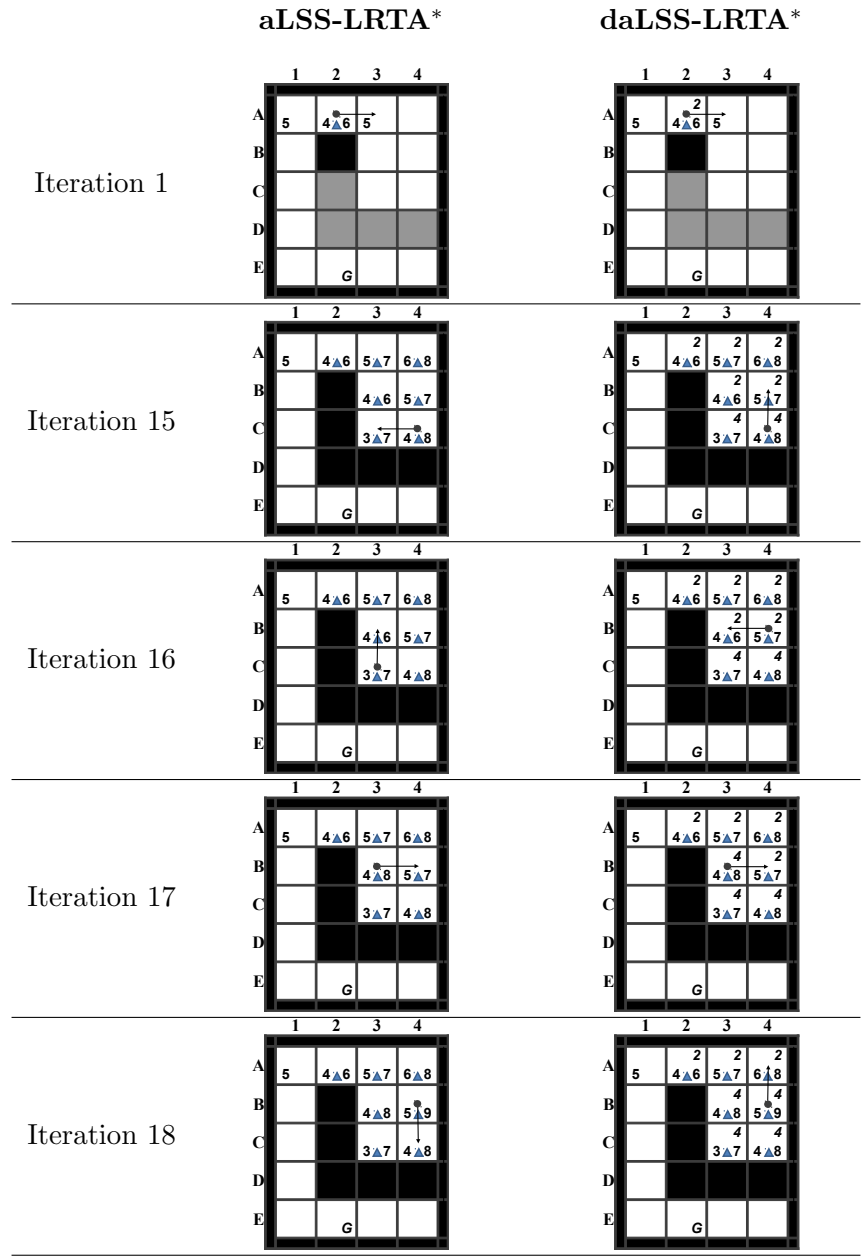

Figure 4: Iterations 1 and 15–18 of aLSS-LRTA* (left) and daLSS-LRTA* (right) with lookahead equal to 1 in a 4-connected grid, analogous to our previous example, in which the objective is the cell E2. In iterations 1 to 14 both algorithms execute in the same way. Numbers in cells correspond to initial $h$-value (lower-left), current $h$-value (lower-right), and difference between those two amounts (upper-right). Triangles (▲) denote states whose heuristic value has been updated.

daLSS-LRTA* and daRTAA* differ, respectively, from LSS-LRTA* and RTAA* in that the selection of the next state to move to (i.e., function `Extract-Best()`) is implemented via Algorithm 9. Note that the worst case complexity of this algorithm is $\mathcal{O}(N \log N)$, where $N$ is the size of $Open$ if binary heaps are used.

---

**Algorithm 9:** Selection of the next state by daRTAA* and daLSS-LRTA*.

**1 function** `Extract-Best-State` $()$
**2** $\quad \Delta_{min} \leftarrow \infty$
**3** $\quad$ **while** $Open \neq \emptyset$ $and$ $\Delta_{min} \neq 0$ **do**
**4** $\quad\quad$ Remove state $s_b$ with smallest $f$-value from $Open$
**5** $\quad\quad$ **if** $h(s_b) - h_0(s_b) < \Delta_{min}$ **then**
**6** $\quad\quad\quad$ $s \leftarrow s_b$
**7** $\quad\quad\quad$ $\Delta_{min} \leftarrow h(s_b) - h_0(s_b)$
**8** $\quad$ **return** $s$

---

Figure 4 illustrates the differences between aLSS-LRTA* and daLSS-LRTA*. Both algorithms execute in the same way if, after the lookahead phase, there is a state in $Open$ whose heuristic value has not been updated. However, when this is not the case (i.e., when the algorithm is "trapped" in a depression), daLSS-LRTA* will move to what seems to be closer to the border of the depression. In the example of Figure 4, at iteration 15, the algorithm chooses B4 instead of C3 since B4 is the state for which the $h$-value has changed the least. After iteration 18, daLSS-LRTA* will move to cells in which less learning has been carried out and thus will exit the depression more quickly.

All the new algorithms presented in this section are closely related. Table 2 shows a schematic view of the different components of each algorithm, and the complexity of the involved algorithms.

## 6. Theoretical Analysis

In this section we analyze the theoretical properties of the algorithms that we propose. We prove that all of our algorithms also satisfy desirable properties that hold for their ancestors. We start off by presenting theoretical results that can be proven using existing proofs available in the literature; among them, we will show that the consistency of the heuristic is maintained by all our algorithms during run time. We continue with results that need different proofs; in particular, termination and convergence to an optimal solution.

As before, we use $h_n$ to refer to the value of variable $h$ at the start of iteration $n$ ($h_0$, thus, denotes the heuristic function given as a parameter to the algorithm). Similarly, $c_n(s, s')$ is the cost of the arc between $s$ and $s'$. Finally, $k_n(s, s')$ denotes the cost of an optimal path between $s$ and $s'$ that traverses only nodes in $Closed$ before ending in $s'$ with respect to cost function $c_n$.

We first establish that if $h$ is initially consistent, then $h$ is non-decreasing over time. This is an important property since it means that the heuristic becomes more accurate over time.

| Algorithm | Update Phase | | Next State Selection | |
|---|---|---|---|---|
| | **Algorithm** | **Time (heaps)** | **Algorithm** | **Time (heaps)** |
| LSS-LRTA* | Modified Dijkstra (Algorithm 4) | $\mathcal{O}(M \log M)$ | Best state in *Open* (Algorithm 3) | $\mathcal{O}(1)$ |
| aLSS-LRTA* | Modified Dijkstra with Marking (Algorithm 6) | $\mathcal{O}(M \log M)$ | Best unmarked state in *Open* (Algorithm 7) | $\mathcal{O}(L \log L)$ |
| daLSS-LRTA* | Modified Dijkstra (Algorithm 4) | $\mathcal{O}(M \log M)$ | State in *Open* that has changed the least (Algorithm 9) | $\mathcal{O}(L \log L)$ |
| RTAA* | Update with best $f$-value (Algorithm 5) | $\mathcal{O}(N)$ | Best state in *Open* (Algorithm 3) | $\mathcal{O}(1)$ |
| aRTAA* | Update with best $f$-value plus marking (Algorithm 8) | $\mathcal{O}(N)$ | Best unmarked state in *Open* (Algorithm 7) | $\mathcal{O}(L \log L)$ |
| daRTAA* | Update with best $f$-value (Algorithm 5) | $\mathcal{O}(N)$ | State in *Open* that has changed the least (Algorithm 9) | $\mathcal{O}(L \log L)$ |

Table 2: Procedures used for the update phase and for the selection of the next state for each of the algorithms discussed in the paper. Worst-case time complexity for each procedure is included assuming the *Open* list is implemented as a binary heap. $M$ corresponds to $|Open| + |Closed|$, $N$ is equal to $|Closed|$, and $L$ is $|Open|$.

**Theorem 1** *If $h_n$ is consistent with respect to cost function $c_n$, then $h_{n+1}(s) \geq h_n(s)$ for any $n$ along an execution of aLSS-LRTA\* or daLSS-LRTA\*.*

**Proof:** Assume the contrary, i.e., that there is a state $s$ such that $h_n(s) > h_{n+1}(s)$. State $s$ must be in *Closed*, since those are the only states whose $h$-value may be updated. As such, by Proposition 1, we have that $h_{n+1}(s) = k_n(s, s_b) + h_n(s_b)$, for some state $s_b$ in *Open*. However, since $h_n(s) > h_{n+1}(s)$, we conclude that:

$$h_n(s) > k_n(s, s_b) + h_n(s_b),$$

which contradicts the fact that $h_n$ is consistent. We thus conclude that the $h$-value of $s$ cannot decrease. □

**Theorem 2** *If $h_n$ is consistent with respect to cost function $c_n$, then $h_{n+1}(s) \geq h_n(s)$ for any $n$ along an execution of aRTAA\* or daRTAA\*.*

**Proof:** Assume the contrary, i.e., that there is a state $s$ such that $h_n(s) > h_{n+1}(s)$. State $s$ must be in *Closed*, since those are the only states whose $h$-value may be updated. The update rule will set the value of $h_{n+1}(s)$ to $f(s') - g(s)$ for some $s' \in Open$, i.e.,

$$h_{n+1}(s) = f(s') - g(s) = g(s') + h_n(s') - g(s).$$

But since $h_n(s) > h_{n+1}(s)$, we have that:

$$h_n(s) > g(s') + h_n(s') - g(s).$$

Reordering terms, we obtain that:

$$h_n(s) + g(s) > g(s') + h_n(s'),$$

which means that the $f$-value of $s$ is greater than the $f$-value of $s'$. It is known however that A\*, run with a consistent heuristic, will expand nodes with non-decreasing $f$-values. We conclude, thus, that $s'$ must have been expanded *before* $s$. Since $s'$ is in *Open*, then $s$ cannot be in *Closed*, which contradicts our initial assumption. We thus conclude that the $h$-value of $s$ cannot decrease. □

**Theorem 3** *If $h_n$ is consistent with respect to cost function $c_n$, then $h_{n+1}$ is consistent with respect to cost function $c_{n+1}$ along an execution aLSS-LRTA\* or daLSS-LRTA\*.*

**Proof:** Since the update procedure used by aLSS-LRTA\*, daLSS-LRTA\* and LSS-LRTA\* update variable $h$ in exactly the same way, the proof by Koenig and Sun (2009) can be reused here. However, we provide a rather simpler proof in Section B.1. □

**Theorem 4** *If $h_n$ is consistent with respect to cost function $c_n$, then $h_{n+1}$ is consistent with respect to cost function $c_{n+1}$ along an execution aRTAA\* or daRTAA\*.*

**Proof:** Since the update procedure used by aRTAA*, daRTAA* and RTAA* update variable $h$ in exactly the same way, we can re-use the proof of Theorem 1 by Koenig and Likhachev (2006b) to establish this result. We provide however a complete proof in Section B.2 □

The objective of the *mark-and-avoid* strategy is to stay away from depressions. The following theorems establish that, indeed, when a state is marked by the aLSS-LRTA* or aRTAA* then such a state is in a heuristic depression of the current heuristic.

**Theorem 5** *Let $s$ be a state such that $s$.updated switches from false to true between iterations $n$ and $n + 1$ in an execution of aLSS-LRTA* or aRTAA* for which $h$ was initially consistent. Then $s$ is in a cost-sensitive heuristic depression of $h_n$.*

**Proof:** We first prove the result for the case of aLSS-LRTA*. The proof for aRTAA* is very similar and can be found in Section B.3.

Let $D$ be the maximal connected component of states connected to $s$ such that:

1. All states in $D$ are in *Closed* after the call to A* in iteration $n$, and

2. Any state $s_d$ in $D$ is such that $h_{n+1}(s_d) > h_n(s_d)$.

Let $s'$ be a state in the boundary of $D$. We first show that $h_n(s') = h_{n+1}(s')$. By definition $s'$ is either in *Closed* or *Open*. If $s' \in Closed$ then, since $s' \notin D$, it must be the case that $s'$ does not satisfy condition 2 of the definition of $D$, and hence $h_{n+1}(s') \leq h_n(s')$. However, since the heuristic is non-decreasing (Theorems 2 and 1), it must be that $h_n(s') = h_{n+1}(s')$. On the other hand, if $s'$ is in *Open*, its heuristic value is not changed and thus also $h_n(s') = h_{n+1}(s')$. We have established, hence, that $h_n(s') = h_{n+1}(s')$.

Now we are ready to establish our result: that $D$ is a cost-sensitive heuristic depression of $h_n$.

Let $s_d$ be a state in $D$. We distinguish two cases.

- Case 1: $s' \in Closed$. Then, by Proposition 1,

$$h_n(s') = k_n(s', s_b) + h_n(s_b), \tag{7}$$

for some $s_b \in Open$. On the other hand, since the heuristic value has increased for $s_d$, $h_n(s_d) < h_{n+1}(s_d) = \min_{s_b' \in Open} k_n(s_d, s_b') + h(s_b')$; in particular, $h_n(s_d) < k_n(s_d, s_b) + h_n(s_b)$. Since $k_n(s_d, s_b)$ is the optimal cost to go from $s_d$ to $s_b$, $k_n(s_d, s_b) \leq k_n(s_d, s') + k_n(s', s_b)$. Substituting $k_n(s_d, s_b)$ in the previous inequality we have:

$$h_n(s_d) < k_n(s_d, s') + k_n(s', s_b) + h_n(s_b). \tag{8}$$

We now substitute the right-hand side of (8) using (7), and we obtain

$$h_n(s_d) < k_n(s_d, s') + h_n(s').$$

- Case 2: $s' \in Open$. Because of Proposition 1 we have $h_{n+1}(s_d) \leq k_n(s_d, s') + h_n(s')$. Moreover, by definition of $D$, we have $h_{n+1}(s_d) > h_n(s_d)$. Combining these two inequalities, we obtain:

$$h_n(s_d) < k_n(s_d, s') + h_n(s').$$

In both cases, we proved $h_n(s_d) < k_n(s_d, s') + h_n(s')$, for any $s_d$ in $D$ and any $s'$ in the boundary of $D$. We conclude $D$ is a cost-sensitive heuristic depression of $h_n$, which finishes the proof. □

Now we turn our attention to termination. We will prove that if a solution exists, then it will be found by any of our algorithms. To prove such a result, we need two intermediate lemmas. The first establishes that when the algorithm moves to the best state in $Open$, then the $h$-value of such a state has not changed more than the $h$-value of the current state. Formally,

**Lemma 1** *Let $s'$ be the state with smallest $f$-value in Open after the lookahead phase of any of aLSS-LRTA\*, daLSS-LRTA\*, aRTAA\*, or daRTAA\*, when initialized with a consistent heuristic $h$. Then,*

$$h_{n+1}(s_{current}) - h_0(s_{current}) \geq h_n(s') - h_0(s').$$

**Proof:** Indeed, by Propositions 1 or 2:

$$h_{n+1}(s_{current}) = k_n(s_{current}, s') + h_n(s') \tag{9}$$

Let $\pi$ be an optimal path found by A\* connecting $s_{current}$ and $s'$. Let $K_0^\pi$ denote the cost of this path with respect to cost function $c_0$. Given that the heuristic $h_0$ is consistent with respect to the graph with cost function $c_0$, we have that $h_0(s_{current}) \leq K_0^\pi + h_0(s')$ which can be re-written as:

$$-h_0(s_{current}) \geq -K_0^\pi - h_0(s'). \tag{10}$$

Adding (10) and (9), we obtain:

$$h_{n+1}(s_{current}) - h_0(s) \geq k_n(s_{current}, s') - K_0^\pi + h_n(s') - h_0(s'). \tag{11}$$

Now, because $c_n$ can only increase, the cost of $\pi$ at iteration $n$, $k_n(s_{current}, s')$, is strictly greater than the cost of $\pi$ at iteration 0, $K_0^\pi$. In other words, the amount $k_n(s_{current}, s') - K_0^\pi$ is positive and can be removed from the right-hand side of (11) to produce:

$$h_{n+1}(s_{current}) - h_0(s) \geq h_n(s') - h_0(s'),$$

which is the desired result. □

The second intermediate result to prove termination is the following lemma.

**Lemma 2** *Let $n$ be an iteration of any of aLSS-LRTA\*, daLSS-LRTA\*, aRTAA\*, or daRTAA\*, when initialized with a consistent heuristic $h$. If $s_{next}$ is not set equal to the state $s'$ with least $f$-value in Open, then:*

$$h_n(s') - h_0(s') > h_n(s_{next}) - h_0(s_{next}).$$

**Proof:** Indeed, if aRTAA\* or aLSS-LRTA\* are run, this means that $s_{next}$ is such that $s_{next}$ is not marked as updated, which means that $h_n(s_{next}) = h_0(s_{next})$, or equivalently, that $h_n(s_{next}) - h_0(s_{next}) = 0$. Moreover, the best state in $Open$, $s'$, was not chosen and hence it must be that $s'$.updated $= true$, which means that $h(s') - h_0(s') > 0$. We obtain then that $h_n(s') - h_0(s') > h_n(s_{next}) - h_0(s_{next})$.

The case of daRTAA* or daLSS-LRTA* is direct by the condition in Line 5 of Algorithm 9. Hence, it is also true that $h_n(s') - h_0(s') > h_n(s_{next}) - h_0(s_{next})$. □

Now we are ready to prove the main termination result.

**Theorem 6** *Let $P$ be an undirected finite real-time search problem such that a solution exists. Let $h$ be a consistent heuristic for $P$. Then, any of aLSS-LRTA\*, daLSS-LRTA\*, aRTAA\*, or daRTAA\*, used with $h$, will find a solution for $P$.*

**Proof:** Let us assume the contrary. There are two cases under which the algorithms do not return a solution: (a) they return "no solution" in Line 5 (Algorithm 1), and (b) the agent traverses an infinite path that never hits a solution node.

For (a) assume any of the algorithms is in state $s$ before the call to A*. When it reaches Line 5 (Algorithm 1), the open list is empty, which means the agent has exhausted the search space of states reachable from $s$ without finding a solution; this is a contradiction with the fact that a solution node is reachable from $s$ and the fact that the search problem is undirected.

For (b) assume that the agent follows an infinite path $\pi$. Observe that in such an infinite execution, after some iteration—say, $R$—the value of variable $c$ does not increase anymore. This is because all states around states in $\pi$ have been observed in the past. As a consequence, in any iteration after $R$ the agent traverses the complete path identified by the A* lookahead procedure (Line 8 in Algorithm 1).

A second important observation is that, after iteration $R$, the value of $h$ for the states in $\pi$ is finite and cannot increase anymore. Indeed, by Theorems 4 and 3, $h$ remains consistent and hence admissible, which means that $h(s)$ is bounded by the actual cost to reach a solution from $s$, for any $s$ in $\pi$. Moreover, since $c$ does not change anymore, the call to the update function will not change the value of $h(s)$, for every $s$ in $\pi$.

Now we are ready to finish the proof. Consider the algorithm executes past iteration $R$. Since the path is infinite and the state space is finite, in some iteration after $R$ the algorithm decides to go back to a previously visited state. As such, we are going to assume the agent visits state $t_0$ and selects to move trough states $t_1 t_2 \cdots t_{r-1} t_r t_0 \cdots$. Since the heuristic does not change anymore, we simply denote it by $h$, regardless of the iteration number. We distinguish two cases.

**Case 1** The agent always decides to move to the best state in *Open*, $s'$, and hence—depending on the algorithm that is used—by Proposition 1 or 2, $h(s) = k(s, s') + h(s')$, which implies $h(s) > h(s')$, since action costs are positive. This implies that:

$$h(t_0) > h(t_1) > h(t_2) > \ldots > h(t_n) > h(t_0),$$

which is a contradiction; it cannot be the case that $h(t_0) > h(t_0)$.

**Case 2** At least once, the agent does not move to the best state in *Open*. Without loss of generality, we assume this happens only once, for a state $t_i$ for some $i < r$. Let $t^*$ be a state with the smallest $f$-value in *Open* after the lookahead is carried out from $t_i$.

By Lemma 1, we can write the following inequalities.

$$h(t_0) - h_0(t_0) \geq h(t_1) - h_0(t_1),$$

$$\vdots$$

$$h(t_{i-1}) - h_0(t_{i-1}) \geq h(t_i) - h_0(t_i),$$
$$h(t_i) - h_0(t_i) \geq h(t^*) - h_0(t^*),$$
$$h(t_{i+1}) - h_0(t_{i+1}) \geq h(t_{i+2}) - h_0(t_{i+2}),$$

$$\vdots$$

$$h(t_r) - h_0(t_r) \geq h(t_0) - h_0(t_0).$$

Let $\mathcal{I}$ be a set containing these inequalities. Now since when in state $t_i$ the algorithm decides to move to $t_{i+1}$ instead of $t^*$, we use Lemma 2 to write:

$$h_n(t^*) - h_0(t^*) > h_n(t_{i+1}) - h_0(t_{i+1}). \tag{12}$$

The inequalities in $\mathcal{I}$ together with (12) entail $h(t_0) - h_0(t_0) > h(t_0) - h_0(t_0)$, which is a contradiction.

In both cases we derive contradictions and hence we conclude the algorithm cannot enter an infinite loop and thus finds a solution. □

We now turn our attention to convergence. The literature often analyzes the properties of real-time heuristic search when they are run on a sequence of *trials* (e.g., Shimbo & Ishida, 2003). Each trial is characterized by running the algorithm from the start state until the problem is solved. The heuristic function $h$ resulting from trial $n$ is used to feed the algorithm's $h$ variable in trial $n + 1$.

Before stating the convergence theorem we prove a result related to how $h$ increases between successive iterations or trials. Indeed, each iteration of our search algorithms potentially increases $h$, making it more informed. The following result implies that this improvement cannot be infinitesimal.

**Lemma 3** *Let $P$ be a finite undirected search problem, and let Sol be a set of states in $P$ from which a solution can be reached. Let $n$ be an iteration of any of aLSS-LRTA\*, daLSS-LRTA\*, aRTAA\*, or daRTAA\*. Then $h_n(s)$ can only take on a finite number of values, for every $s$ in $P$.*

**Proof:** Given Proposition 1, along an execution of any of the algorithms of the LSS-LRTA\* family, it is simple to prove by induction on $n$ that:

$$h_n(s) = K + h_0(s'''),$$

for any $n$, where $K$ is sum of the costs of 0 or more arcs in $P$ under cost function $c_n$.

On the other hand, given the update rule of any of the algorithms of the RTAA\* family (e.g., Line 6 in Algorithm 8),

$$h_n(s) = K - K' + h_0(s'''),$$

for any $n$, where $K$ and $K'$ correspond to the sum of the costs of some arcs in $P$ under cost function $c_n$.

Since in finite problems there is a finite number of arcs, the quantities referred to by $K$ and $K'$ can only take on a finite number of values. This implies that $h_n(s)$, for any $s$ in $P$, can only take on a finite number of values, which concludes the proof. □

Below we show that if $h$ converges after a sequence of trials, the solution found with $h$ is optimal.

**Theorem 7** *Let $P$ be an undirected finite real-time search problem such that a solution exists. Let $h$ be a consistent heuristic for $P$. When initialized with $h$, a sequence of trials of any of aLSS-LRTA\*, daLSS-LRTA\*, aRTAA\*, or daRTAA\*, converges to an optimal solution.*

**Proof:** First, observe that since the heuristic is admissible, it remains admissible after a number of trials are run. This is a consequence of Theorems 3 and 4. Hence, for every state $s$ from which a goal state can be reached, $h(s)$ is bounded from above by the (finite amount) $h^*(s)$.

On the other hand, by Lemma 3, the $h$-values of states from which a solution is reachable can only increase a finite number of times. After a sequence of trials the value of $h$ thus converges; i.e., at least for one complete trial, $h(s)$ is not changed, for every $s$ in $P$. We can also assume that in such a trial, the value of $c$ does not change either, since once $h$ converges, the same path of states is always followed and thus no new cost increases are made.

Let us focus on a run of any of our algorithms in which both $h$ and $c$ do not change. Observe that this means that $h_n(s) = h_0(s)$ for any $n$ (recall $h_0$ is the heuristic given as input to the algorithm). Independent of the algorithm used, this implies the algorithm always moves to the best state in *Open*. Let $s_1 \ldots s_m$ be the sequence of states that were assigned to $s_{next}$ during the execution ($s_m$ is thus a goal state). Observe that since $c$ does not change along the execution, states $s_1 \ldots s_m$ are actually visited by the agent. Depending on the algorithm that is used, by Proposition 1 or 2, we know:

$$h(s_i) = k(s_i, s_{i+1}) + h(s_{i+1}), \qquad \text{for all } i \in \{0, \ldots, m-1\}, \tag{13}$$

where $k(s_i, s_{i+1})$ is the cost of an optimal path between $s_i$ and $s_{i+1}$. Since the heuristic is consistent $h(s_m) = 0$, and thus with the family of equations in (13) we conclude $h(s_0)$ is equal to $\sum_{i=0}^{m-1} k(s_i, s_{i+1})$, which corresponds to the cost of the path traversed by the agent. But we know that $h$ is also admissible, so:

$$h(s_0) = \sum_{i=0}^{m-1} k(s_i, s_{i+1}) \leq h^*(s_0).$$

Since $h^*(s_0)$ *is* the cost of an optimal solution, we conclude the path found has an optimal cost. □

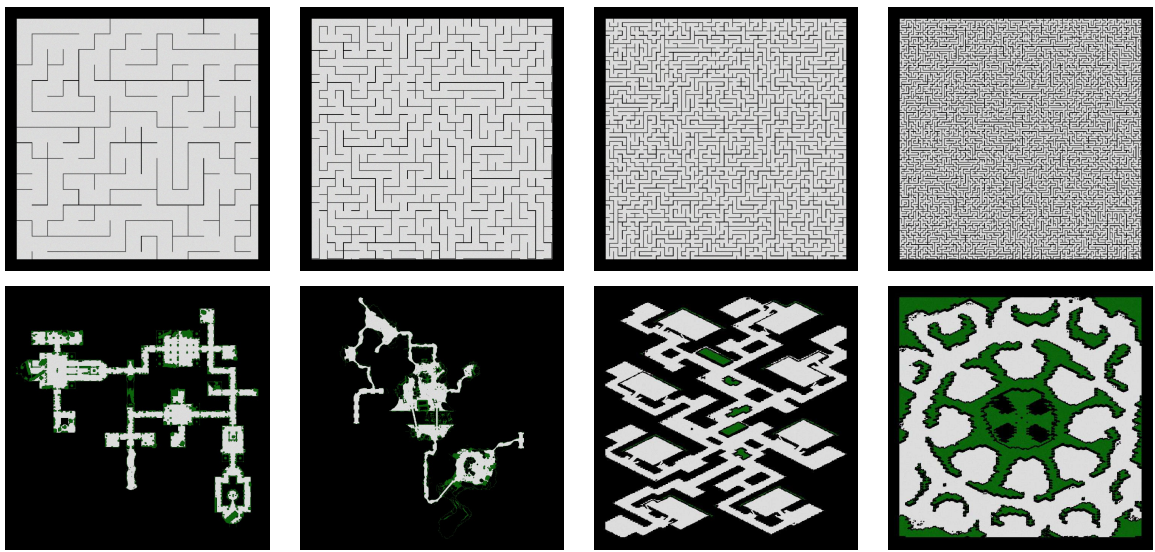

Figure 5: Upper row: the four maze maps used to test our approach; each of $512 \times 512$ cells. Lower row: 4 out of the 12 game maps used. The first two come from *Dragon Age: Origins*; the remaining 2 are from *StarCraft*.

## 7. Empirical Evaluation

We evaluated our algorithms at solving real-time navigation problems in unknown environments. LSS-LRTA* and RTAA* are used as a baseline for our comparisons. For fairness, we used comparable implementations that use the same underlying codebase. For example, all search algorithms use the same implementation for binary heaps as priority queues and break ties among cells with the same $f$-values in favor of cells with larger $g$-values, which is known to be a good tie-breaking strategy.

We carried out our experiments over two sets of benchmarks: deployed game maps and mazes. We used twelve maps from deployed video games to carry out the experiments. The first six are taken from the game *Dragon Age*, and the remaining six are taken from the game *StarCraft*. The maps were retrieved from Nathan Sturtevant's pathfinding repository.[1] In addition, we used four maze maps taken from the HOG2 repository.[2] They are shown in Figure 5. All results were obtained using a Linux machine with an Intel Xeon CPU running at 2GHz and 12 GB RAM.

All maps are regarded as undirected, eight-neighbor grids. Horizontal and vertical movements have cost 1, whereas diagonal movements have cost $\sqrt{2}$. We used the *octile distance* (Sturtevant & Buro, 2005) as heuristic.

---

1. `http://www.movingai.com/` and `http://hog2.googlecode.com/svn/trunk/maps/`. For Dragon Age we used the maps brc202d, orz702d, orz900d, ost000a, ost000t and ost100d of size $481 \times 530$, $939 \times 718$, $656 \times 1491$ $969 \times 487$, $971 \times 487$, and $1025 \times 1024$ cells respectively. For StarCraft, we used the maps ArcticStation, Enigma, Inferno JungleSiege, Ramparts and WheelofWar of size $768 \times 768$, $768 \times 768$, $768 \times 768$, $768 \times 768$, $512 \times 512$ and $768 \times 768$ cells respectively.

2. `http://hog2.googlecode.com/svn/trunk/maps/`

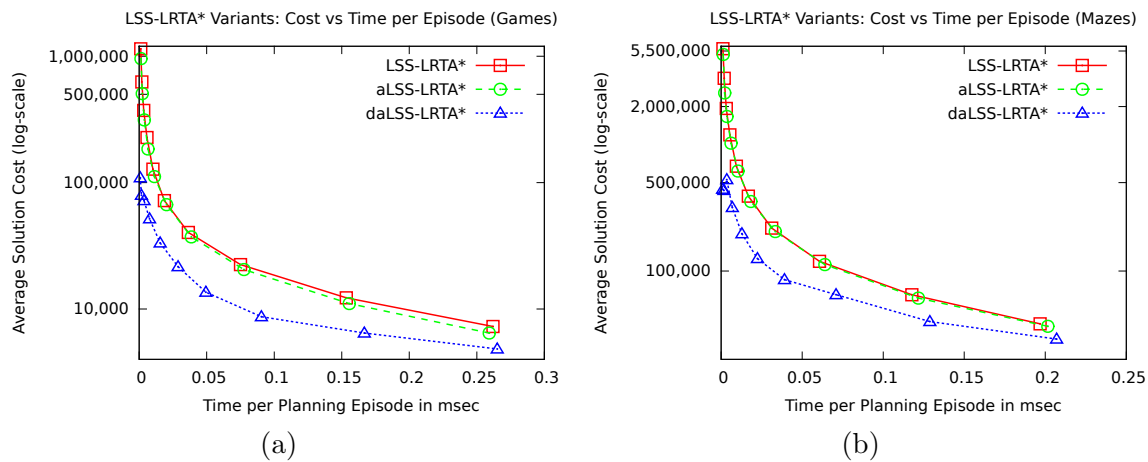

Figure 6: Plots showing the average solution cost found by the LSS-LRTA* variants versus average planning time per episode, measured in milliseconds. (a) shows stats on the game-map benchmarks, and (b) on the mazes benchmarks. Times are shown in milliseconds. Costs are shown on a log-scale.

For our evaluation we ran all algorithms for 10 different lookahead values. For each map, we generate 500 test cases. For each test case we choose the start and goal cells randomly.

In the presentation of our results we sometimes use the concept of *improvement factor*. When we say that the improvement factor of an algorithm $A$ with respect to $B$ in terms of average solution cost is $n$, it means that on average $A$ produces solutions that are $n$ times cheaper than the ones found by $B$.

Next we describe the different views of the experimental data that is shown in plots and tables. We then continue to draw our experimental conclusions.

## 7.1 An Analysis of the LSS-LRTA* Variants

This section analyzes the performance of LSS-LRTA*, aLSS-LRTA* and daLSS-LRTA*. Figure 6 shows two plots for the average solution costs versus the average planning time per episode for the three algorithms in games and mazes benchmarks. Planning time per planning episode is an accurate measure of the effort carried out by each of the algorithms. Thus these plots illustrate how solution quality varies depending on the effort that each algorithm carries out.

Regardless of the search effort, we observe aLSS-LRTA* slightly but consistently outperforms LSS-LRTA* in solution cost. In games benchmarks we observe that for equal search effort, aLSS-LRTA* produces average improvement factors between 1.08 and 1.20 in terms of solution cost. In mazes, on the other hand, improvement factors are between 1.04 and 1.25. In games, the largest improvements are observed when the lookahead parameter (and hence the search time per episode) is rather small. Thus aLSS-LRTA*'s advantage over LSS-LRTA* is more clearly observed when tighter time constraints are imposed on planning episodes.

Often times results in real-time search literature are presented in the form of tables, with search performance statistics reported per each lookahead value. We provide such tables the appendix of the paper (Tables 5 and 6). An important observation that can be drawn from the tables is that time per planning episode in LSS-LRTA* and aLSS-LRTA* are very similar for a fixed lookahead value; indeed, the time per planning episode of aLSS-LRTA* is only slightly larger than that of LSS-LRTA*. This is interesting since it shows that the worst-case asymptotic complexity does not seem to be achieved for aLSS-LRTA* (cf. Table 2).

The experimental results show that daLSS-LRTA*'s more refined mechanism for escaping depressions is better than that of aLSS-LRTA*. For any given value of the search effort, daLSS-LRTA* consistently outperforms aLSS-LRTA* by a significant margin in solution cost in games and mazes. daLSS-LRTA* also outperforms aLSS-LRTA* in total search time, i.e., the overall time spent searching until a solution is found. Details can be found in Tables 5 and 6. When the search effort for each algorithm is small, daLSS-LRTA*'s average solution quality is substantially better than aLSS-LRTA*'s; the improvements are actually close to an order of magnitude.

daLSS-LRTA* consistently outperforms LSS-LRTA* by a significant margin in total search time and solution quality, independent of the search effort employed. In terms of solution cost daLSS-LRTA* produces average improvement factors with respect to LSS-LRTA* between 1.66 and an order of magnitude in the game benchmarks, and produces average improvement factors between 1.49 and an order of magnitude in the mazes benchmarks. For a fixed lookahead (see Tables 5 and 6 for the specific numbers), the time spent per planning episode by daLSS-LRTA* is larger than time spent per planning episode by LSS-LRTA* because daLSS-LRTA* makes more heap percolations than LSS-LRTA*. However, for small values of the lookahead parameter, daLSS-LRTA* obtains better solutions using less time per planning episode than LSS-LRTA* used with a much larger lookahead. For example, in game maps, with a lookahead parameter equal to 32, daLSS-LRTA* obtains better solutions than LSS-LRTA* with the lookahead parameter equal to 128, requiring, on average, 2.6 times *less time* per planning episode. In mazes, with a lookahead parameter equal to 16, daLSS-LRTA* obtains better solutions than LSS-LRTA* with the lookahead parameter equal to 64, requiring, on average, 2.4 times *less time* per planning episode.

For low values of the lookahead parameter (i.e. very limited search effort) daLSS-LRTA* obtains better solutions in less time per planning episode than aLSS-LRTA* used with a much larger lookahead. For example, in game maps, with a lookahead parameter equal to 1, daLSS-LRTA* obtains better solutions than aLSS-LRTA* with the lookahead parameter equal to 16, requiring, on average, 14.1 times *less time* per planning episode. On the other hand, in mazes with a lookahead parameter equal to 1, daLSS-LRTA* obtains better solutions than aLSS-LRTA* with the lookahead parameter equal to 16, requiring, on average, 11.6 times less time per planning episode.

For a fixed lookahead (see Tables 5 and 6), the time taken by daLSS-LRTA* per planning episode is larger than the time taken by aLSS-LRTA* per planning episode. This increase can be explained because, on average, daLSS-LRTA*'s open list grows larger than that of aLSS-LRTA*. This is due to the fact that, in the benchmarks we tried, daLSS-LRTA* tends to expand cells that have less obstacles around than aLSS-LRTA* does. As a result, daLSS-LRTA* expands more cells in the learning phase or makes more heap percolations in the lookahead phase than aLSS-LRTA*.

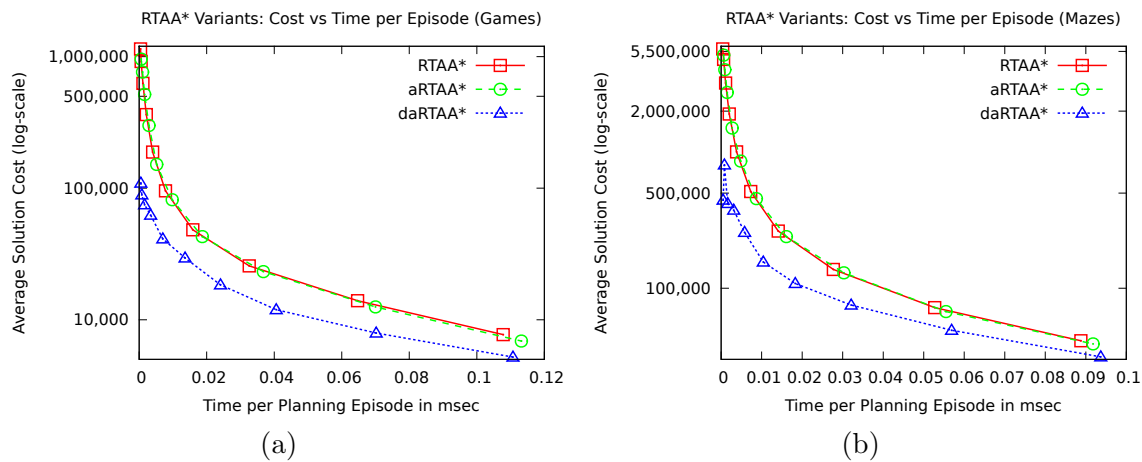

(a)  (b)

Figure 7: Plots showing the average solution cost found by the RTAA* variants versus average planning time per episode. (a) shows stats on the game-maps benchmarks, and (b) on the mazes benchmarks. Costs are shown on a log-scale.

Results show that, among the LSS-LRTA* variants, daLSS-LRTA* is the algorithm with the best performance. In fact daLSS-LRTA* is clearly superior to LSS-LRTA*. Of the 60,000 runs (12 maps × 500 test cases × 10 lookahead-values) in game benchmarks, daLSS-LRTA* obtains a better solution quality than LSS-LRTA* in 69.9% of the cases, they tie in 20.9% of the cases, and LSS-LRTA* obtains a better-quality solution in only 9.2% of the cases.

Of the 20,000 (4 maps × 500 test cases × 10 lookahead-values) runs in mazes benchmarks, daLSS-LRTA* obtains a better solution quality than LSS-LRTA* in 75.1% of the cases, they tie in 3.3% of the cases, and LSS-LRTA* obtains a better-quality solution in 21.7% of the cases.

## 7.2 An Analysis of the RTAA* Variants

In this section we analyze the relative performance of RTAA*, aRTAA*, and daRTAA*. Figure 7 shows two plots of the average solution costs versus the average effort carried out per search episode.

For the same search effort, we do not observe significant improvements of aRTAA* over RTAA*. Indeed, only for small values of the average time per search episode does aRTAA* improve the solution quality upon that of RTAA*. In general, however, both algorithms seem to have very similar performance.

On the other hand, the results show that daRTAA*'s mechanism for escaping depressions is substantially better than that of aRTAA*. For small values for the lookahead parameter (and hence reduced search effort), daRTAA* obtains better solutions than the other variants used with a much larger lookahead. Indeed, for limited search effort, daRTAA* is approximately an order of magnitude better than the two other algorithms. For example, in game maps, with a lookahead parameter equal to 1, daRTAA* obtains better solutions than aRTAA* with the lookahead parameter equal to 16, requiring, on average, 10.4 times less time per planning episode.

daRTAA* substantially improves RTAA*, which is among the best real-time heuristic search algorithms known to date. In game maps, daRTAA* needs only a lookahead parameter of 16 to obtain solutions better than RTAA* with the lookahead parameter of 64. With those values, daRTAA* requires about 2.3 times less time per planning episode than RTAA*.

Our results show that daRTAA* is the best-performing algorithm of the RTAA* family. Of the 60,000 runs in game-map benchmarks, daRTAA* obtains a better solution quality than RTAA* in 71.2% of the cases, they tie in 20.5% of the cases, and RTAA* obtains a better-quality solution in only 8.3% of the cases. Of the 20,000 runs in mazes, daRTAA* obtains a better solution quality than RTAA* in 78.0% of the cases, they tie in 2.7% of the cases, and RTAA* obtains a better-quality solution in 19.4% of the cases.

### 7.3 daLSS-LRTA* **Versus** daRTAA*

daRTAA*, the best performing algorithm among the RTAA* variants, is also superior to daLSS-LRTA*, the best-performing algorithm of the LSS-LRTA* variants. Figure 8 shows average solution costs versus search effort, in game maps and mazes.

As can be seen in the figure, when the lookahead parameter is small (i.e., search effort is little), the performance of daRTAA* and daLSS-LRTA* is fairly similar. However, as more search is allowed per planning episode, daRTAA* outperforms daLSS-LRTA*. For example, in games benchmarks, daRTAA*, when allowed to spend 0.08 milliseconds per episode, will obtain solutions comparable to those of daLSS-LRTA* but when allowed do spend 0.18 millisecconds per episode.

Furthermore, the slopes of the curves are significantly more favorable to daRTAA* over daLSS-LRTA*. This can be verified in both types of benchmarks and is important since it speaks to an inherent superiority of the RTAA* framework when time per planning episode is the most relevant factor.

### 7.4 An Analysis of Disaggregated Data

The performance of real-time algorithms usually varies depending on the map used. To illustrate how the algorithms perform in different maps, Figure 9 shows the improvement on solution cost of daLSS-LRTA* over LSS-LRTA* on 4 game and 4 maze benchmarks. They confirm that improvements can be observed in all domains thus showing that average values are representative of daLSS-LRTA*'s behavior in individual benchmarks. Although aLSS-LRTA* and daLSS-LRTA* outperform LSS-LRTA* on average, there are specific cases in which the situation does not hold. Most notably, we observe that in one of the maze benchmarks daLSS-LRTA* does not improve significantly with respect to LSS-LRTA* for large values of the lookahead parameter. We discuss this further in the next section. Figure 10 shows also the improvement factors of daRTAA* over RTAA*. In this plot, the different algorithms show a similar relative performance in relation to the LSS-LRTA* variants.

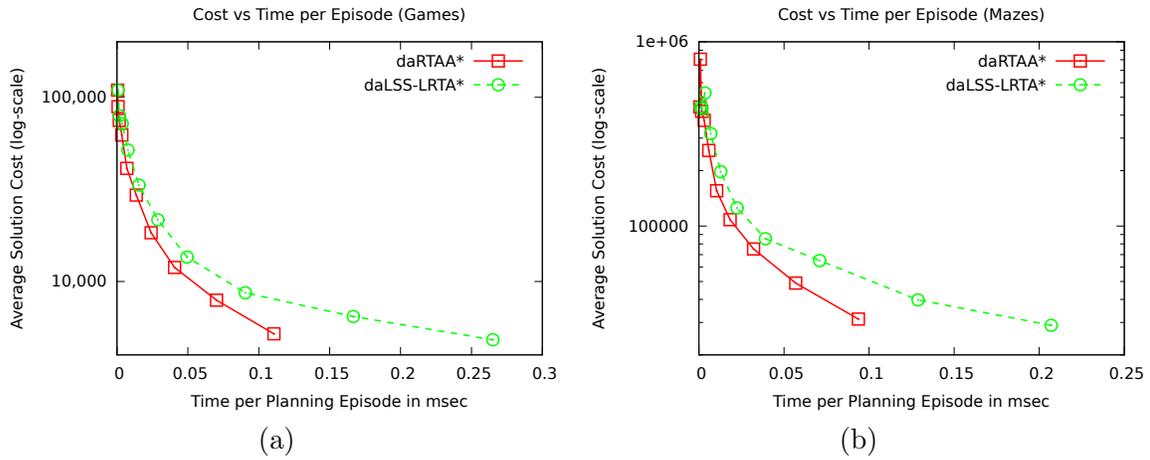

Figure 8: Plots showing the average solution cost found by the daRTAA* and daLSS-LRTA* versus average planning time per episode. (a) shows stats on the game-maps benchmarks, and (b) on the mazes benchmarks. Costs are shown on a log-scale.

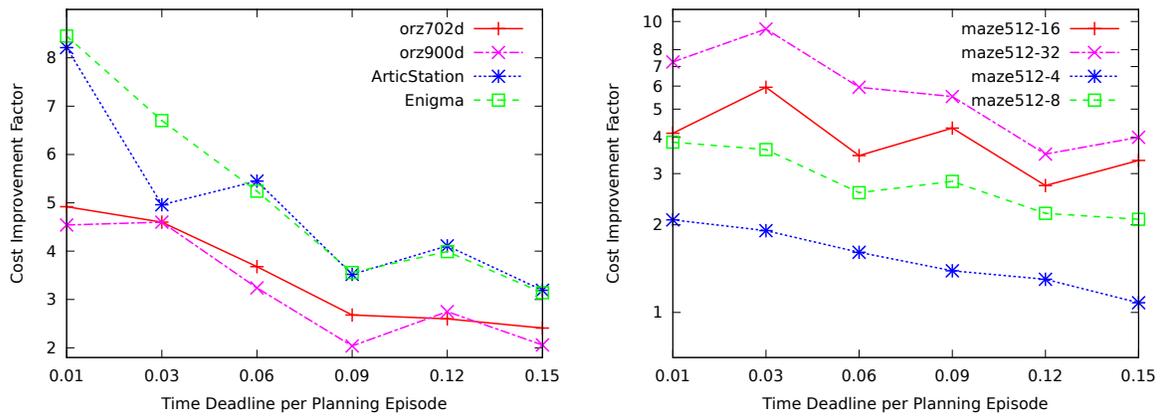

Figure 9: Cost improvement factor of daLSS-LRTA* over LSS-LRTA*, in game maps (left) and maze benchmarks (right). An improvement factor equal to $n$ indicates that the solution found by our algorithm is $n$ times cheaper than the one found by the original algorithm.

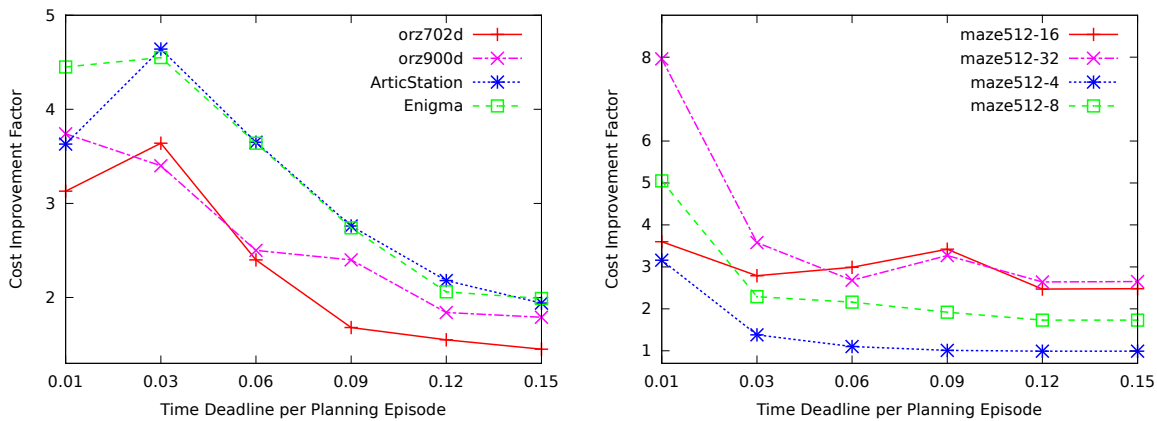

Figure 10: Cost improvement factor of daRTAA* over RTAA*, in game maps (left) and maze (right) benchmarks. An improvement factor equal to $n$ indicates that the solution found by our algorithm is $n$ times cheaper than the one found by the original algorithm.

## 7.5 A Worst-Case Experimental Analysis

Although all our algorithms perform a resource-bounded computation per planning episode, it is hard to tune the lookahead parameter in such a way that both LRTA* and daLSS-LRTA* will incur the same worst-case planning effort. This is because the time spent in extracting the best state from the open list depends on the structure of the search space expanded in each lookahead phase.

In this section we set out to carry an experimental worst-case analysis based on a theoretical worst-case bound. This bound is obtained from the worst-case effort per planning step as follows. If RTAA* performs $k$ expansions per planning episode, then the open list could contain up to $8k$ states. This is because each state has at most 8 neighbors. In the worst case, the effort spent in adding all such states to the open list would be $8k \log 8k$. On the other hand, daRTAA* would make the same effort to insert those states into the open list, but would incur an additional cost of $8k \log 8k$, in the worst-case, to remove all states from the open list. Therefore, in a worst-case scenario, given a lookahead parameter equal to $k$, daRTAA* will make double the effort than RTAA* makes for the same parameter.

Based on that worst-case estimation, Figure 11 presents the performance of the RTAA* variants, displacing the RTAA* curve by a lookahead factor of 2. We conclude that in this worst-case scenario daRTAA* still clearly outperforms RTAA*. Gains vary from one order of magnitude, for low values of the lookahead parameter, to very similar performance when the lookahead parameter is high.

We remark, however, that we never observed this worst-case in practice. For example, in our game benchmarks, RTAA*, when used with a lookahead parameter $2k$ spends, on average 50% *more* time per planning episode than daRTAA* used with lookahead parameter $k$.

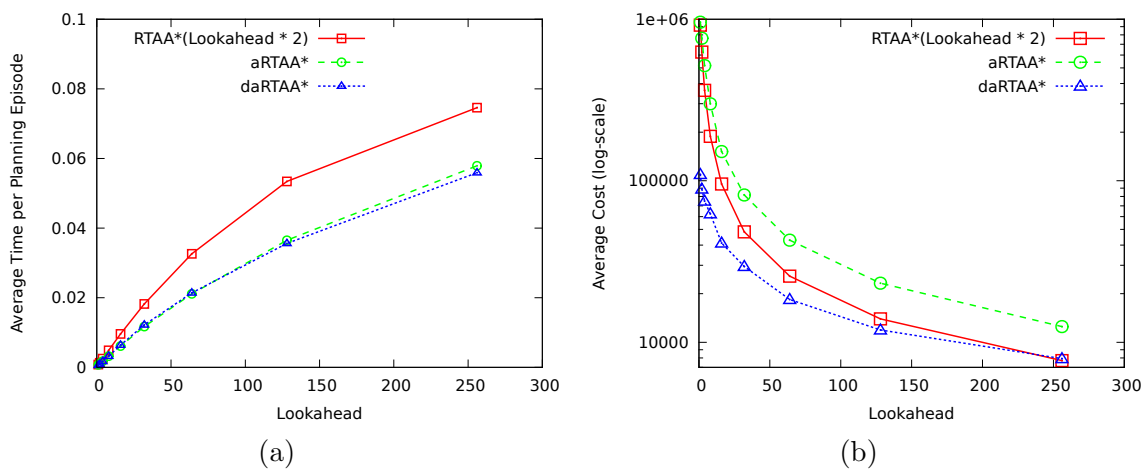

Figure 11: Plots showing the average time per planning episode and average solution cost per lookahead parameter, adjusting the performance of RTAA* using a theoretical worst-case bound of 2. As such, for RTAA*, the average cost reported for for a lookahead of $k$ actually corresponds to the cost obtained for a lookahead $2k$. Costs are shown on a log-scale.

## 8. Discussion

There are a number of aspects of our work that deserve a discussion. We focus on two of them. First, we discuss the setting in which we have evaluated our work, which focused on showing performance improvements in the *first trial* for a search in an a priori *unknown domain*, without considering other settings. Second, we discuss in which scenarios our algorithms may not exhibit average performance improvements that were shown in the previous section.

### 8.1 The Experimental Setting: Unknown Environments, First Trial

Our algorithm is tailored to solving quickly a search problem in which the environment is initially unknown. This setting has several applications, including goal-directed navigation in unknown terrain (Koenig et al., 2003; Bulitko & Lee, 2006). It has also been widely used to evaluate real-time heuristic search algorithms (e.g., Koenig, 1998; Hernández & Meseguer, 2005; Bulitko & Lee, 2006; Hernández & Meseguer, 2007; Koenig & Sun, 2009).

On the other hand, we did not present an evaluation of our algorithm in environments that are *known* a priori. In a previous paper (Hernández & Baier, 2011d), however, we showed that aLSS-LRTA* obtains similar improvements over LSS-LRTA* when the environment is known. However, we omit results on known environments since RTAA* and LSS-LRTA* are not representative of the state of the art in those scenarios. Indeed, algorithms like TBA* (Björnsson et al., 2009) outperform LSS-LRTA* significantly. It is not immediately obvious how to incorporate our techniques to algorithms like TBA*.

We did not present experimental results regarding convergence after several successive search trials. Recall that in this setting, the agent is "teleported" to the initial location and a new search trial is carried out. Most real-time search algorithms—ours included—are guaranteed to eventually find an optimal solution. Our algorithms do not particularly excel in this setting. This is because the heuristic value of fewer states is updated, and hence the heuristic values for states in the search space converges slowly to the correct value. As such, generally more trials are needed to converge.

Convergence performance is important for problems that are solved *offline* and in which real-time approaches may be more adequate for computing an approximation of the optimal solution. This is the case of the problem of computing an optimal policy in MDPs using Real-Time Dynamic Programming (Barto, Bradtke, & Singh, 1995). We are not aware, however, of any application in deterministic search in which searching offline using real-time search would yield better performance than using other suboptimal search algorithms (e.g., Richter, Thayer, & Ruml, 2010; Thayer, Dionne, & Ruml, 2011). Indeed, Wilt, Thayer, and Ruml (2010) concluded that real-time algorithms, though applicable, should not be used for solving shortest path problems unless there is a need for real-time action.

## 8.2 Bad Performance Scenarios

Although our algorithms clearly outperform its originators LSS-LRTA* and RTAA* on average, it is possible to contrive families of increasingly difficult path-finding tasks in which our algorithms perform worse than their respective predecessors.

Consider for example the 4-connected grid-world scenario of size $7 \times n$ shown in Figure 12. The goal of the agent is to reach the state labeled with $G$, starting from $S$. Assume furthermore that to solve this problem we run aRTAA* or aLSS-LRTA*, with lookahead parameter equal to 1, and that ties are broken such that the *up* movement has priority over the *down* movement. In the initial state both algorithms will determine that the initial state (cell E3) is in a heuristic depression and thus will update the heuristic of cell E3. Cell E3 is now marked as in a depression. Since both cells D3 and F3 have the same heuristic value and ties are broken in favor of upper cells, the agent is then moved to cell D3. In later iterations, the algorithm will not prefer to move to cells that have been updated and therefore the agent will not go back to state E3 unless it is currently in D3 and (at least) C3 is also marked. However, the agent will not go back to D3 quickly. Indeed, it will visit *all states* to the right of *Wall 1* and *Wall 2* before coming back to E3. This happens because, as the algorithm executes, it will update and mark all visited states, and will never prefer to go back to a previously marked position unless *all* current neighbors are also marked.

In the same situation, RTAA* and LSS-LRTA*, run with lookahead parameter 1 will behave differently depending on the tie-breaking rules. Indeed, if the priority is given by *up* (highest), *down*, *right*, and *left* (lowest), then both RTAA* and LSS-LRTA* find the goal fairly quickly as they do not have to visit states to the right of the walls. Indeed, since the tie-breaking rules prefer a move up, the agent reaches cell A3 after 4 moves, and then proceeds straight to the goal. In such situations, the performance of aRTAA* or aLSS-LRTA* can be made arbitrarily *worse* than that of RTAA* or LSS-LRTA*, as $n$ is increased.

A quite different situation is produced if the tie-breaking follows the priorities given by *up* (highest), *right*, *down*, and *left* (lowest). In this case all four algorithms have to visit the states to the right of both walls. Indeed, once A3 is reached, there is a tie between the $h$-value of B3 and A4. The agent prefers moving to A4, and from there on it continues moving to the right of the grid in a zig-zag fashion.

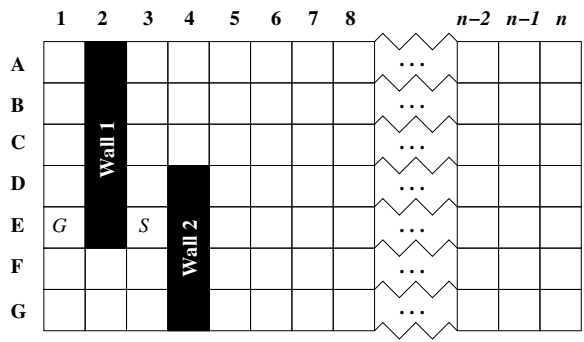

Figure 12: A situation in which the relative performance between LSS-LRTA* and aLSS-LRTA* changes depending on the value of $n$. $S$ is the start state, and $G$ is the goal. Ties are broken in favor of upper cells.

After investigating executions of our "da-" algorithms in the maze512-4-0 benchmark (performance is shown in Figures 9 and 10), we believe that the lack of improvement in this particular benchmark can be explained by the situation just described. This benchmark is a $512 \times 512$ maze in which corridors have a 4-cell width. For low lookahead values, the number of updates is not high enough to "block" the corridors. As such, for low values of the lookahead parameter the increase in performance is still reasonably good. As the lookahead increases, the algorithm updates more states in one single iteration, and, as a result, chances are that good paths may become blocked.

Interestingly, however, we do not observe this phenomenon on mazes with wider corridors or on game maps. A necessary condition to "block" a corridor that leads to a solution is that the agent has sufficient knowledge about the borders of the corridor. In mazes with narrow corridors this may happen with relative ease, as the agent only needs a few moves to travel between opposite walls. In grids in which corridors are wide however, knowledge about the existence of obstacles (walls) is hard to obtain by the agent, and, thus, the chances of updating *and* blocking, a corridor that leads to a solution are lower.

We believe that it is possible to *prove* that our algorithms are always better or always worse for specific search space topologies. We think, nevertheless, that such an analysis may be hard to carry out, and that its practical significance may be limited. Therefore we decided to exclude it from the scope of this work. On the other hand, we think that the impressive performance exhibited by our algorithms in many benchmarks is sufficiently strong in favor of using our algorithms in domains that do not contain narrow corridors.

## 9. Related Work

Besides LSS-LRTA* and RTAA*, there are a number of real-time search algorithms that can be used in a priori unknown environments. LRTA*$(k)$ and LRTA*$_{\mathrm{LS}}(k)$ (Hernández & Meseguer, 2005, 2007) are two algorithms competitive with LSS-LRTA* that are capable of learning the heuristic of several states at the same time; the states for which the heuristic is learned is independent from those expanded in the lookahead phase. They may escape heuristic depressions more quickly than LRTA*, but its action selection mechanism is not aware of heuristic depressions. eLSS-LRTA* is a preliminary version of aLSS-LRTA* we presented in an extended abstract (Hernández & Baier, 2011a). It is outperformed by aLSS-LRTA* on average, as it usually becomes too focused on avoiding depressions.

Our algorithms have been designed in order to find good-quality solutions on the first search trial. Other algorithms described in the literature have been designed with different objectives in mind. For example, RIBS (Sturtevant, Bulitko, & Björnsson, 2010) is a real-time algorithm specifically designed to converge quickly to an optimal solution. It will move the agent as if an iterative-deepening A* search was carried out. As such the first solution it finds is optimal. As a consequence, RIBS potentially requires more time to find one solution than LSS-LRTA* does, but if an optimal solution is required RIBS will likely outperform LSS-LRTA* run to convergence. $f$-LRTA* (Sturtevant & Bulitko, 2011) is another recent real-time search algorithm which builds upon ideas introduced by RIBS, in which the $g$-cost of states is learned through successive trials. It has good convergence performance, but needs to do more computation per planning step than LSS-LRTA*.

Incremental A* methods, like D* (Stentz, 1995), D*Lite (Koenig & Likhachev, 2002), Adaptive A* (Koenig & Likhachev, 2006a), and Tree Adaptive A* (Hernández, Sun, Koenig, & Meseguer, 2011), are search methods that also allow solving goal-directed navigation problems in unknown environments. If the first-move delay is required to be short, incremental A* methods cannot be used since they require to compute a complete solution before starting to move. Real-time search remains the only applicable strategy for this task when limited time is allowed per planning episode.

Less related to our work are algorithms that abide to real-time search constraints but that assume the environment is known in advance and that sufficient time is given prior to solving the problem, allowing *preprocessing*. Examples are D LRTA* (Bulitko, Luštrek, Schaeffer, Björnsson, & Sigmundarson, 2008) and kNN-LRTA* (Bulitko et al., 2010), tree subgoaling (Hernández & Baier, 2011b), or real-time search via compressed path databases (Botea, 2011).

Finally, the concept of cost-sensitive depression in real-time search could be linked to other concepts used to describe the poor performance of planning algorithms. For example, Hoffmann (2005, 2011) analyzed the existence of plateaus in $h^+$, an effective admissible domain-independent planning heuristic, and how this negatively affects the performance of otherwise fast planning algorithms. Cushing, Benton, and Kambhampati (2011) introduced the concept of $\varepsilon$-traps that is related to poor performance of best-first search in problems in which action costs have a high variance. $\varepsilon$-traps are areas of the search space connected by actions of least cost. As such, the $h$-values of states in $\varepsilon$-traps is not considered in their analysis. Although we think that the existence of cost-sensitive heuristic depressions does affect the performance of A*, the exact relation between the performance of A* and heuristic depressions does not seem to be obvious.

## 10. Summary and Future Work

We have presented a simple principle for guiding real-time search algorithms away from heuristic depressions. We proposed two alternative approaches for implementing the principle: *mark-and-avoid* and *move-to-border*. In the first approach, states that are proven to be in a depression are marked in the update phase, and then avoided, if possible, when deciding the next move. In the second approach, the algorithm selects as the next move the state that seems closer to the border of a depression.

Both approaches can be implemented efficiently. Mark-and-avoid requires very little overhead, which results in an almost negligible increment in time per planning episode. Move-to-border, on the other hand, requires more overhead per planning episode, but, given a time deadline per planning episode, it is able to obtain the best-quality solutions.

Experimentally, we have shown that in goal-directed navigation tasks in unknown terrain, our algorithms outperform their predecessors RTAA* and LSS-LRTA*. Indeed, the algorithms based on move-to-border—daLSS-LRTA* and daRTAA*—are significantly more efficient than LSS-LRTA* and RTAA*, especially when the lookahead parameter is a small value.

The four algorithms proposed have good properties: in undirected, finite search spaces, they are guaranteed to find a solution if such a solution exists. Moreover, they converge to an optimal solution after running a number of search trials.

Depression avoidance is a principle applicable to other real-time heuristic search algorithms. Indeed, we think it could be easily incorporated into LRTA*$(k)$, LRTA*$_{LS}(k)$, and P-LRTA* (Rayner et al., 2007). All those algorithms have specialized mechanisms for updating the heuristic, but the mechanism to select the next state is just like LSS-LRTA*'s run with lookahead parameter equal to 1. We think significant improvements could be achieved if the procedure to select the next movement was changed by daLSS-LRTA*'s. We also believe depression avoidance could be incorporated into multi-agent real-time search algorithms (e.g., Knight, 1993; Yokoo & Kitamura, 1996; Kitamura, Teranishi, & Tatsumi, 1996).

## 11. Acknowledgments

We thank the JAIR reviewers who provided extensive feedback that helped improve this article significantly. We also thank the IJCAI-11, SoCS-11, and AIIDE-11 anonymous reviewers for their thoughtful insights on earlier versions of this work. We are very grateful to Cristhian Aguilera, who helped out running some of the experiments. Carlos Hernández was partly funded a by Fondecyt project #11080063. Jorge Baier was partly funded by the VRI-38-2010 grant from Pontificia Universidad Católica de Chile and the Fondecyt grant number 11110321.

## Appendix A. Additional Experimental Data

Tables 3–6 show average statistics for LSS-LRTA*, RTAA*, and our 4 algorithms.

**RTAA***

| LookAhead parameter | Avg. Cost | # Planning Episodes | Total Time | Time per Episode | Exp. per ep. | Perc. per ep. |
|---|---|---|---|---|---|---|
| 1 | 5,731,135 | 5,307,571 | 2,174 | 0.0004 | 1.0 | 5.9 |
| 2 | 4,805,384 | 3,885,684 | 2,410 | 0.0006 | 2.0 | 9.1 |
| 4 | 3,217,283 | 2,147,060 | 2,321 | 0.0011 | 4.0 | 16.8 |
| 8 | 1,905,895 | 954,571 | 1,912 | 0.0020 | 8.0 | 33.2 |
| 16 | 1,004,971 | 353,707 | 1,338 | 0.0038 | 16.0 | 67.8 |
| 32 | 513,692 | 127,555 | 927 | 0.0073 | 32.0 | 145.6 |
| 64 | 262,760 | 46,777 | 661 | 0.0141 | 63.9 | 331.8 |
| 128 | 137,403 | 18,787 | 521 | 0.0277 | 126.3 | 816.5 |
| 256 | 71,939 | 9,012 | 475 | 0.0527 | 237.8 | 2,016.6 |
| 512 | 41,089 | 5,973 | 530 | 0.0888 | 397.4 | 4,101.6 |

**aRTAA***

| LookAhead parameter | Avg. Cost | # Planning Episodes | Total Time | Time per ep. | Exp. per ep. | Perc. per ep. |
|---|---|---|---|---|---|---|
| 1 | 5,165,062 | 4,785,257 | 2,798 | 0.0006 | 1.0 | 10.6 |
| 2 | 4,038,347 | 3,260,134 | 2,981 | 0.0009 | 2.0 | 18.9 |
| 4 | 2,746,638 | 1,832,375 | 2,829 | 0.0015 | 4.0 | 34.0 |
| 8 | 1,504,379 | 755,334 | 2,034 | 0.0027 | 8.0 | 61.5 |
| 16 | 859,669 | 305,372 | 1,458 | 0.0048 | 16.0 | 113.3 |
| 32 | 455,023 | 114,089 | 992 | 0.0087 | 32.0 | 216.0 |
| 64 | 239,484 | 43,497 | 699 | 0.0161 | 63.9 | 440.6 |
| 128 | 129,765 | 18,478 | 559 | 0.0303 | 126.3 | 988.9 |
| 256 | 67,346 | 9,108 | 506 | 0.0555 | 237.5 | 2,272.5 |
| 512 | 38,939 | 6,172 | 567 | 0.0918 | 399.4 | 4,394.9 |

daRTAA*

| $k$ | Avg. Cost | # Planning Episodes | Total Time | Time per ep. | Exp. per ep. | Perc. per ep. |
|---|---|---|---|---|---|---|
| 1 | 443,773 | 415,327 | 208 | 0.0005 | 1.0 | 10.5 |
| 2 | 804,990 | 689,014 | 575 | 0.0008 | 2.0 | 27.3 |
| 4 | 419,616 | 321,418 | 502 | 0.0016 | 4.0 | 58.5 |
| 8 | 374,684 | 260,163 | 801 | 0.0031 | 8.0 | 129.4 |
| 16 | 257,126 | 148,616 | 864 | 0.0058 | 16.0 | 261.1 |
| 32 | 155,573 | 66,818 | 697 | 0.0104 | 32.0 | 476.7 |
| 64 | 108,337 | 34,119 | 626 | 0.0183 | 63.8 | 854.8 |
| 128 | 75,158 | 17,686 | 568 | 0.0321 | 126.2 | 1,536.7 |
| 256 | 49,065 | 10,370 | 590 | 0.0569 | 239.3 | 2,920.9 |
| 512 | 31,265 | 6,954 | 652 | 0.0937 | 408.9 | 5,074.1 |

Table 3: Average results of RTAA* variants over mazes. For a given lookahead parameter value, we report the average solution cost (Avg. Cost), average number of planning episodes (# Planning Episodes), total runtime (Total Time), average runtime per search episode (Time per Episode), average number of expansions per episode (Exp. per ep.), average number of percolations per planning episode (Perc. per ep.). All times are reported in milliseconds.

**RTAA\***

| LookAhead parameter | Avg. Cost | # Planning Episodes | Total Time | Time per Episode | Exp. per ep. | Perc. per ep. |
|---|---|---|---|---|---|---|
| 1 | 1,146,014 | 1,058,265 | 448 | 0.0004 | 1.0 | 6.1 |
| 2 | 919,410 | 747,824 | 475 | 0.0006 | 2.0 | 9.4 |
| 4 | 626,623 | 422,389 | 469 | 0.0011 | 4.0 | 17.3 |
| 8 | 363,109 | 184,753 | 384 | 0.0021 | 8.0 | 34.1 |
| 16 | 188,346 | 67,652 | 269 | 0.0040 | 16.0 | 70.1 |
| 32 | 95,494 | 24,609 | 193 | 0.0078 | 32.0 | 152.9 |
| 64 | 48,268 | 9,138 | 146 | 0.0159 | 63.9 | 361.3 |
| 128 | 25,682 | 3,854 | 126 | 0.0326 | 126.4 | 932.3 |
| 256 | 13,962 | 1,941 | 126 | 0.0647 | 236.8 | 2,351.8 |
| 512 | 7,704 | 1,220 | 132 | 0.1078 | 377.6 | 4,616.7 |

**aRTAA\***

| LookAhead parameter | Avg. Cost | # Planning Episodes | Total Time | Time per ep. | Exp. per ep. | Perc. per ep. |
|---|---|---|---|---|---|---|
| 1 | 958,795 | 885,506 | 549 | 0.0006 | 1.0 | 11.1 |
| 2 | 763,367 | 621,438 | 598 | 0.0010 | 2.0 | 19.9 |
| 4 | 516,545 | 348,785 | 569 | 0.0016 | 4.0 | 36.0 |
| 8 | 299,786 | 154,037 | 445 | 0.0029 | 8.0 | 66.1 |
| 16 | 151,737 | 55,706 | 290 | 0.0052 | 16.0 | 122.5 |
| 32 | 81,695 | 21,533 | 210 | 0.0098 | 32.0 | 235.0 |
| 64 | 42,883 | 8,357 | 157 | 0.0187 | 63.9 | 485.5 |
| 128 | 23,217 | 3,631 | 134 | 0.0368 | 126.3 | 1,114.1 |
| 256 | 12,510 | 1,845 | 129 | 0.0700 | 235.8 | 2,586.7 |
| 512 | 6,892 | 1,178 | 133 | 0.1132 | 372.7 | 4,826.3 |

**daRTAA\***

| $k$ | Avg. Cost | # Planning Episodes | Total Time | Time per ep. | Exp. per ep. | Perc. per ep. |
|---|---|---|---|---|---|---|
| 1 | 109,337 | 102,616 | 53 | 0.0005 | 1.0 | 11.6 |
| 2 | 88,947 | 79,951 | 66 | 0.0008 | 2.0 | 29.6 |
| 4 | 74,869 | 62,664 | 103 | 0.0016 | 4.0 | 68.0 |
| 8 | 62,400 | 48,838 | 165 | 0.0034 | 8.0 | 153.7 |
| 16 | 41,145 | 28,453 | 199 | 0.0070 | 16.0 | 327.4 |
| 32 | 29,469 | 16,857 | 229 | 0.0136 | 32.0 | 654.9 |
| 64 | 18,405 | 8,152 | 196 | 0.0241 | 63.9 | 1,167.4 |
| 128 | 11,924 | 3,908 | 158 | 0.0406 | 126.4 | 1,958.9 |
| 256 | 7,921 | 2,116 | 149 | 0.0702 | 238.3 | 3,491.5 |
| 512 | 5,205 | 1,311 | 145 | 0.1107 | 385.1 | 5,654.4 |

Table 4: Average results of RTAA* variants over game maps. For a given lookahead parameter value, we report the average solution cost (Avg. Cost), average number of planning episodes (# Planning Episodes), total runtime (Total Time), average runtime per search episode (Time per Episode), average number of expansions per episode (Exp. per ep.), average number of percolations per planning episode (Perc. per ep.). All times are reported in milliseconds.

**LSS-LRTA***

| LookAhead parameter | Avg. Cost | # Planning Episodes | Total Time | Time per Episode | Exp. per ep. | Perc. per ep. |
|---|---|---|---|---|---|---|
| 1 | 5,731,135 | 5,307,571 | 6,036 | 0.0011 | 8.5 | 14.3 |
| 2 | 3,346,675 | 2,594,738 | 4,967 | 0.0019 | 13.4 | 28.0 |
| 4 | 1,931,251 | 1,247,205 | 4,009 | 0.0032 | 20.7 | 52.1 |
| 8 | 1,195,330 | 586,084 | 3,187 | 0.0054 | 32.9 | 97.6 |
| 16 | 674,872 | 233,400 | 2,189 | 0.0094 | 54.5 | 182.9 |
| 32 | 391,120 | 96,163 | 1,613 | 0.0168 | 95.2 | 367.4 |
| 64 | 218,303 | 39,002 | 1,215 | 0.0312 | 175.6 | 799.4 |
| 128 | 119,177 | 16,649 | 1,010 | 0.0607 | 341.3 | 1,939.0 |
| 256 | 64,861 | 8,420 | 991 | 0.1177 | 655.0 | 4,704.4 |
| 512 | 38,182 | 5,805 | 1,143 | 0.1968 | 1,079.2 | 8,961.1 |

**aLSS-LRTA***

| LookAhead parameter | Avg. Cost | # Planning Episodes | Total Time | Time per ep. | Exp. per ep. | Perc. per ep. |
|---|---|---|---|---|---|---|
| 1 | 5,165,062 | 4,785,257 | 6,174 | 0.0013 | 8.5 | 19.0 |
| 2 | 2,561,769 | 1,981,509 | 4,321 | 0.0022 | 13.3 | 37.7 |
| 4 | 1,670,535 | 1,078,512 | 3,923 | 0.0036 | 20.7 | 69.5 |
| 8 | 1,027,134 | 504,696 | 3,069 | 0.0061 | 33.0 | 126.6 |
| 16 | 617,302 | 213,959 | 2,217 | 0.0104 | 54.6 | 228.3 |
| 32 | 354,691 | 87,700 | 1,603 | 0.0183 | 95.8 | 441.1 |
| 64 | 205,214 | 37,106 | 1,240 | 0.0334 | 176.9 | 918.4 |
| 128 | 112,288 | 16,069 | 1,028 | 0.0640 | 344.4 | 2,134.7 |
| 256 | 61,031 | 8,300 | 1,010 | 0.1217 | 659.1 | 4,997.9 |
| 512 | 36,524 | 5,879 | 1,185 | 0.2016 | 1,082.0 | 9,283.8 |

**daLSS-LRTA***

| LookAhead parameter | Avg. Cost | # Planning Episodes | Total Time | Time per ep. | Exp. per ep. | Perc. per ep. |
|---|---|---|---|---|---|---|
| 1 | 443,773 | 415,327 | 357 | 0.0009 | 6.2 | 18.9 |
| 2 | 433,576 | 353,087 | 603 | 0.0017 | 11.7 | 43.3 |
| 4 | 527,638 | 393,222 | 1,374 | 0.0035 | 40.0 | 96.6 |
| 8 | 317,508 | 205,868 | 1,412 | 0.0069 | 40.0 | 225.8 |
| 16 | 197,066 | 100,984 | 1,293 | 0.0128 | 70.7 | 459.9 |
| 32 | 125,511 | 45,682 | 1,023 | 0.0224 | 119.7 | 816.1 |
| 64 | 85,373 | 22,725 | 888 | 0.0391 | 209.7 | 1,477.1 |
| 128 | 65,009 | 13,772 | 977 | 0.0709 | 384.7 | 2,936.7 |
| 256 | 39,777 | 8,201 | 1,056 | 0.1288 | 698.4 | 5,972.6 |
| 512 | 28,937 | 6,330 | 1,310 | 0.2070 | 1,115.8 | 10,136.2 |

Table 5: Average results of LSS-LRTA* variants over mazes. For a given lookahead parameter value, we report the average solution cost (Avg. Cost), average number of planning episodes (# Planning Episodes), total runtime (Total Time), average runtime per search episode (Time per Episode), average number of expansions per episode (Exp. per ep.), average number of percolations per planning episode (Perc. per ep.). All times are reported in milliseconds. Results obtained over a Linux PC with a Pentium QuadCore 2.33 GHz CPU and 8 GB RAM.

**LSS-LRTA\***

| LookAhead parameter | Avg. Cost | # Planning Episodes | Total Time | Time per Episode | Exp. per ep. | Perc. per ep. |
|---|---|---|---|---|---|---|
| 1 | 1,146,014 | 1,058,265 | 1,260 | 0.0012 | 8.7 | 14.8 |
| 2 | 625,693 | 488,096 | 979 | 0.0020 | 13.7 | 29.3 |
| 4 | 372,456 | 242,171 | 818 | 0.0034 | 21.3 | 54.3 |
| 8 | 227,526 | 113,236 | 654 | 0.0058 | 33.8 | 102.4 |
| 16 | 127,753 | 45,242 | 460 | 0.0102 | 56.1 | 193.5 |
| 32 | 72,044 | 18,445 | 345 | 0.0187 | 98.7 | 397.7 |
| 64 | 40,359 | 7,687 | 280 | 0.0364 | 184.9 | 903.4 |
| 128 | 22,471 | 3,444 | 258 | 0.0750 | 370.1 | 2,338.1 |
| 256 | 12,264 | 1,774 | 272 | 0.1534 | 733.6 | 6,003.8 |
| 512 | 7,275 | 1,192 | 312 | 0.2620 | 1,207.5 | 11,548.9 |

**aLSS-LRTA\***

| LookAhead parameter | Avg. Cost | # Planning Episodes | Total Time | Time per ep. | Exp. per ep. | Perc. per ep. |
|---|---|---|---|---|---|---|
| 1 | 958,795 | 885,506 | 1,185 | 0.0013 | 8.7 | 19.8 |
| 2 | 506,745 | 395,546 | 903 | 0.0023 | 13.7 | 40.1 |
| 4 | 313,789 | 204,478 | 786 | 0.0038 | 21.3 | 73.7 |
| 8 | 184,632 | 92,594 | 602 | 0.0065 | 34.1 | 135.6 |
| 16 | 111,633 | 39,857 | 449 | 0.0113 | 56.6 | 246.0 |
| 32 | 66,911 | 17,271 | 351 | 0.0203 | 99.5 | 479.6 |
| 64 | 37,215 | 7,217 | 278 | 0.0386 | 186.8 | 1,036.2 |
| 128 | 20,524 | 3,234 | 251 | 0.0776 | 374.5 | 2,553.8 |
| 256 | 11,053 | 1,677 | 261 | 0.1556 | 741.4 | 6,339.3 |
| 512 | 6,460 | 1,137 | 295 | 0.2592 | 1,204.5 | 11,823.7 |

**daLSS-LRTA\***

| $k$ | Avg. Cost | # Planning Episodes | Total Time | Time per ep. | Exp. per ep. | Perc. per ep. |
|---|---|---|---|---|---|---|
| 1 | 109,337 | 102,616 | 86 | 0.0008 | 6.1 | 20.8 |
| 2 | 79,417 | 69,976 | 116 | 0.0017 | 12.1 | 49.9 |
| 4 | 72,028 | 58,931 | 214 | 0.0036 | 23.3 | 118.2 |
| 8 | 51,753 | 38,862 | 300 | 0.0077 | 44.1 | 274.0 |
| 16 | 33,351 | 20,792 | 322 | 0.0155 | 80.3 | 586.2 |
| 32 | 21,622 | 10,177 | 293 | 0.0288 | 139.6 | 1,122.4 |
| 64 | 13,581 | 4,715 | 233 | 0.0494 | 236.9 | 1,911.2 |
| 128 | 8,693 | 2,424 | 220 | 0.0905 | 435.4 | 3,725.4 |
| 256 | 6,464 | 1,604 | 267 | 0.1667 | 791.5 | 7,538.1 |
| 512 | 4,830 | 1,195 | 317 | 0.2651 | 1,237.3 | 12,697.3 |

Table 6: Average results of LSS-LRTA\* variants over game maps. For a given lookahead parameter value, we report the average solution cost (Avg. Cost), average number of planning episodes (# Planning Episodes), total runtime (Total Time), average runtime per search episode (Time per Episode), average number of expansions per episode (Exp. per ep.), average number of percolations per planning episode (Perc. per ep.). All times are reported in milliseconds. Results obtained over a Linux PC with a Pentium QuadCore 2.33 GHz CPU and 8 GB RAM.

# Appendix B. Additional Proofs for Theorems

## B.1 Proof of Theorem 3

We establish that, for any pair of neighbor states, $s$ and $s'$, $h_{n+1}(s) \leq c_{n+1}(s, s') + h_{n+1}(s')$. We divide the rest of the argument in three cases.

***Case 1***. Both $s$ and $s'$ are in *Closed*. Then, by Proposition 1,

$$h_{n+1}(s') = k_n(s', s'') + h_n(s''), \tag{14}$$

for some $s'' \in Open$. On the other hand, again by Proposition 1,

$$h_{n+1}(s) = \min_{s_b \in Open} k_n(s, s_b) + h_n(s_b),$$

and thus

$$h_{n+1}(s) \leq k_n(s, s'') + h_n(s''), \tag{15}$$

since $s''$ is an element of *Open*. However, because $k_n(s, s'')$ is the cost of the shortest path between $s$ and $s''$, we know that

$$k_n(s, s'') \leq c_n(s, s') + k_n(s', s'') \tag{16}$$

Adding up (15) and (16), we obtain

$$h_{n+1}(s) \leq c_n(s, s') + k_n(s', s'') + h_n(s'') \tag{17}$$

Using Equation 14 we substitute $k_n(s', s'') + h_n(s'')$ in Inequality 17, obtaining:

$$h_{n+1}(s) \leq c_n(s, s') + h_n(s'). \tag{18}$$

Since the cost function can only increase, we have that $c_n(s, s') \leq c_{n+1}(s, s')$, and hence:

$$h_{n+1}(s) \leq c_{n+1}(s, s') + h_n(s'), \tag{19}$$

Finally, since $h$ is non-decreasing (Theorem 1), we have $h_n(s') \leq h_{n+1}(s')$, which allows us to write

$$h_{n+1}(s) \leq c_{n+1}(s, s') + h_{n+1}(s'), \tag{20}$$

which finishes the proof for this case.

***Case 2***. One state among $s$ and $s'$ is in *Closed*, and the other state is not in *Closed*. Without loss of generality, assume $s \in Closed$. Since $s'$ is not in *Closed*, it must be in *Open*, because $s$ was expanded by A* and $s'$ is a neighbor of $s$. By Proposition 1 we know:

$$h_{n+1}(s) = \min_{s_b \in Open} k_n(s, s_b) + h_n(s_b),$$

but since $s'$ is a particular state in *Open*, we have:

$$h_{n+1}(s) \leq c_n(s, s') + h_n(s').$$

Since $c_n \leq c_{n+1}$, we obtain:

$$h_{n+1}(s) \leq c_{n+1}(s, s') + h_n(s'),$$

which concludes the proof for this case.

***Case 3***. Both $s$ and $s'$ are not in *Closed*. Since $h_n$ is consistent:

$$h_n(s) \leq c_n(s, s') + h_n(s') \tag{21}$$

Now we use that the $h$-value of $s$ and $s'$ are not updated ($h_n(s) = h_{n+1}(s)$ and $h_n(s') = h_{n+1}(s')$), and the fact that the cost function increases to write:

$$h_{n+1}(s) \leq c_{n+1}(s, s') + h_{n+1}(s'), \tag{22}$$

which finishes the proof for this case.

In all three cases we proved the desired inequality and therefore we conclude the heuristic $h_{n+1}$ is consistent with respect to cost function $c_{n+1}$.

## B.2 Proof of Theorem 4

We establish that, for any pair of neighbor states, $s$ and $s'$, $h_{n+1}(s) \leq c_{n+1}(s, s') + h_{n+1}(s')$. We divide the rest of the argument in three cases.

***Case 1***. Both $s$ and $s'$ are in *Closed*. We have that

$$h_{n+1}(s) = f(s^*) - g(s), \tag{23}$$
$$h_{n+1}(s') = f(s^*) - g(s'), \tag{24}$$

for some $s^*$ in *Open*. Subtracting (24) from (23), we obtain:

$$h_{n+1}(s) - h_{n+1}(s') = g(s') - g(s). \tag{25}$$

Since $h_n$ is consistent $g(s)$ and $g(s')$ correspond to the cost of the shortest path between $s_{current}$ and, respectively, $s$ and $s'$. Thus $g(s') = k_n(s_{current}, s')$ and $g(s) = k_n(s_{current}, s)$, and therefore:

$$h_{n+1}(s) - h_{n+1}(s') = k_n(s_{current}, s') - k_n(s_{current}, s). \tag{26}$$

Let us consider a path from $s_{current}$ to $s'$ that goes optimally to $s$, and then goes from $s$ to $s'$. The cost of such a path must be at least $k_n(s_{current}, s')$. In other words:

$$k_n(s_{current}, s') \leq k_n(s_{current}, s) + c_n(s, s'),$$

which directly implies:

$$k_n(s_{current}, s') - k_n(s_{current}, s) \leq c_n(s, s'). \tag{27}$$

Now we combine (27) and (26) to obtain:

$$h_{n+1}(s) \leq c_n(s, s') + h_{n+1}(s'). \tag{28}$$

And, finally, since $c_n \leq c_{n+1}$ we conclude that:

$$h_{n+1}(s) \leq c_{n+1}(s, s') + h_{n+1}(s'), \tag{29}$$

which finishes the proof for this case.

**Case 2.** One state among $s$ and $s'$ is in *Closed*, and the other state is not in *Closed*. Without loss of generality, assume $s' \in Closed$. Since $s'$ is not in *Closed*, it must be in *Open*, because $s$ was expanded by A$^*$ and $s'$ is a neighbor of $s$.

For some state $s^*$ in *Open*, we have that

$$h_{n+1}(s) = f(s^*) - g(s) \tag{30}$$

Again we use the fact that, with the consistent heuristic $h_n$, A$^*$ expands nodes with increasing $f$-values. Note that $s^*$ is the state that would have been expanded next by A$^*$, and that $s'$ would have been expanded later on. Moreover, as soon as $s'$ would have been expanded the $g$-value for $s'$ is the optimal cost of the path from $s_{current}$ to $s'$, $k_n(s_{current}, s')$. Therefore, we can write:

$$f(s^*) \leq k_n(s_{current}, s') + h_n(s'), \tag{31}$$

as $k_n(s_{current}, s') + h_n(s')$ is the $f$-value of $s'$ upon expansion. Adding up (30) and (31), we obtain:

$$h_{n+1}(s) \leq k_n(s_{current}, s') - g(s) + h_n(s')$$

However, since $s$ is in *Closed*, $g(s)$ is the cost of an optimal path from $s_{current}$ to $s$, and thus:

$$h_{n+1}(s) \leq k_n(s_{current}, s') - k_n(s_{current}, s) + h_n(s') \tag{32}$$

We use now the same argument of the previous case to conclude that:

$$k_n(s_{current}, s') - k_n(s_{current}, s) \leq c_n(s, s'). \tag{33}$$

Combining (31) and (33) we obtain:

$$h_{n+1}(s) \leq c_n(s, s') + h_n(s') \tag{34}$$

Since $s'$ is not in closed, $h_{n+1}(s') = h_n(s)$. Furthermore, we know that $c_n \leq c_{n+1}$. Substituting in (34), we obtain:

$$h_{n+1}(s) \leq c_{n+1}(s, s') + h_{n+1}(s'), \tag{35}$$

which allows us to conclude the proof for this case.

**Case 3.** Both $s$ and $s'$ are not in *Closed*. The proof is the same as that for Case 3 in Theorem 3.

In all three cases we proved the desired inequality and therefore we conclude the heuristic $h_{n+1}$ is consistent with respect to cost function $c_{n+1}$.

## B.3 An Appendix for the Proof of Theorem 5

This section describes the proof of Theorem 5 for the specific case of aRTAA$^*$.

Let $D$ be the maximal connected component of states connected to $s$ such that (1) all states in $D$ are in *Closed* after the call to A$^*$ in iteration $n$, and (2) any state $s_d$ in $D$ is such that $h_{n+1}(s_d) > h_n(s_d)$. We prove that $D$ is a cost-sensitive heuristic depression of $h_n$.

Let $s'$ be a state in the boundary of $D$; as argued for the case of aLSS-LRTA$^*$, we can show that $h_n(s') = h_{n+1}(s')$. Now, let $s_d$ be a state in $D$. We continue the proof by showing that $h_n(s_d)$ is too low with respect to $h_n(s')$, which means that $D$ is a heuristic depression of $h_n$. For this final part of the proof, we distinguish two cases: (Case 1) $s' \in Closed$, and (Case 2) $s' \in Open$.

For Case 1, given that $h_{n+1}(s') = h_n(s')$, we know $h_n(s') = f^* - g(s')$, where $f^*$ is the lowest $f$-value in the open list after the algorithm is run, and hence:

$$f^* = h_n(s') + g(s') \tag{36}$$

On the other hand, since by definition of $D$ the heuristic value has increased for $s_d$,

$$h_n(s_d) < h_{n+1}(s_d) = f^* - g(s_d). \tag{37}$$

Substituting $f^*$ in Eq. 37 with the right-hand-side of Eq. 36, we get:

$$h_n(s_d) < h_n(s') + g(s') - g(s_d). \tag{38}$$

Because the heuristic is consistent and both $s'$ and $s_d$ are in $Closed$, $g(s')$ and $g(s_d)$ actually correspond to the cost of the cheapest path to reach, respectively, $s'$ and $s_d$ from $s$; i.e., $g(s') = k(s, s')$ and $g(s_d) = k_n(s, s_d)$. In addition, the triangular inequality $k_n(s, s_d) + k_n(s_d, s') \geq k_n(s, s')$, can be re-written as:

$$g(s') - g(s_d) \leq k_n(s_d, s'). \tag{39}$$

Inequalities 38 and 39 imply $h_n(s_d) < k_n(s_d, s') + h_n(s')$.

Finally, for Case 2, if $s' \in Open$, by Proposition 2 and the fact that $h_{n+1}(s_d) > h_n(s_d)$, we also have that $h_n(s_d) < k_n(s_d, s') + h_n(s')$.

In both cases, we proved $h_n(s_d) < k_n(s_d, s') + h_n(s')$, for any $s_d$ in $D$ and any $s'$ in the boundary of $D$. We conclude $D$ is a cost-sensitive heuristic depression of $h_n$.